\RequirePackage{fix-cm}
\documentclass[smallextended]{svjour3}       
\smartqed  
\usepackage{graphicx}
\usepackage{subfigure}
 \usepackage{epstopdf}
 \usepackage{epsfig}
 \DeclareGraphicsExtensions{.pdf, .jpeg, .png}
\usepackage[colorlinks,linkcolor=red]{hyperref}  
\usepackage{bigstrut,multirow,rotating}  
\usepackage{booktabs}                  
\usepackage{enumerate}
\usepackage{amsmath}
\usepackage{bbm}
\usepackage{amsfonts,amssymb}
\usepackage{appendix}
\usepackage[numbers,sort&compress]{natbib}
\usepackage{url}
\usepackage[hyphenbreaks]{breakurl}

\begin{document}

\title{An evolutionary approach for discovering non-Gaussian stochastic dynamical systems based on nonlocal Kramers-Moyal formulas
}

\titlerunning{An evolutionary approach for discovering non-Gaussian systems}        

\author{Yang Li        \and
        Shengyuan Xu    \and
        Jinqiao Duan$^{\ast}$
}

\institute{Li Y. and Xu S. \at
               School of Automation, Nanjing University of Science and Technology, 200 Xiaolingwei Street, Nanjing 210094, China           
              \and
              Duan J.$^{\ast}$ \at
              Department of Mathematics and Department of Physics, Great Bay University, Dongguan, Guangdong 523000, China
              \\
              \\
              $^{\ast}$Corresponding author:
              \email{duan@gbu.edu.cn}
}

\date{Received: date / Accepted: date}

\maketitle

\begin{abstract}
Discovering explicit governing equations of stochastic dynamical systems with both (Gaussian) Brownian noise and (non-Gaussian) L\'evy noise from data is chanllenging due to possible intricate functional forms and the inherent complexity of L\'evy motion. This present research endeavors to develop an evolutionary symbol sparse regression (ESSR) approach to extract non-Gaussian stochastic dynamical systems from sample path data, based on nonlocal Kramers-Moyal formulas, genetic programming, and sparse regression. More specifically, the genetic programming is employed to generate a diverse array of candidate functions, the sparse regression technique aims at learning the coefficients associated with these candidates, and the nonlocal Kramers-Moyal formulas serve as the foundation for constructing the fitness measure in genetic programming and the loss function in sparse regression. The efficacy and capabilities of this approach are showcased through its application to several illustrative models. This approach stands out as a potent instrument for deciphering non-Gaussian stochastic dynamics from available datasets, indicating a wide range of applications across different fields.

\keywords{Nonlocal Kramers-Moyal formulas \and stochastic dynamical systems \and evolutionary symbolic sparse regression method \and L\'evy motion \and genetic programming}
\subclass{MSC 60F10 \and MSC 60J25 \and MSC 65J60}
\end{abstract}

\section{Introduction}
\label{intro}

The ubiquity of non-Gaussian noise in various fields, from finance to physics and biology, has led to a surge of interest in comprehending its influence on dynamical systems under random fluctuations. Non-Gaussian noise, characterized by its heavy-tailed distributions and potential for describing extreme events, is pivotal in driving complex dynamics that traditional Gaussian models fail to capture \cite{Duan2015}. Its applications range from financial market volatility modeling \cite{tankov2003financial, rachev2011financial}, to the examination of turbulent flows in fluid dynamics \cite{shlesinger1987levy}, and even to the switching phenomena in gene transcriptional regulatory systems \cite{xu2013levy}, where the departure from Gaussianity can lead to significant deviations in system behavior. Additionally, Ditlevsen \cite{Ditlevsen} demonstrated that the climate change in earth follows a stochastic system with $\alpha$-stable L\'evy noise based on real observational data of Greenland ice core. Kanazawa et al. \cite{kanazawa2020loopy} confirmed through experimental and theoretical research that the motion of particles in liquids obeys a non-Gaussian L\'evy distribution under the influence of planktonic microorganisms. Consequently, stochastic systems driven by non-Gaussian noise serve as phenomenological models with profound scientific significance, capable of describing many complex dynamical behaviors that Gaussian distributions cannot characterize. The need to accurately represent and predict the impact of non-Gaussian noise has become increasingly apparent, motivating the development of advanced modeling techniques that can encapsulate its distinctive properties.

Owing to the limited comprehension of intricate systems, such as those in neuroscience and climate dynamics, traditional modeling approaches that rely on first principles often encounter significant challenges. Fortunately, the rapid advancement of observational tools and computational technologies has facilitated the emergence of numerous data-driven modeling techniques to extract the governing equations of complex systems in recent years. By learning directly from data, these approaches can uncover intricate relationships and patterns that might be obscured in traditional model-based analyses. For example, the sparse regression method was proposed to discover deterministic ordinary \cite{SINDy1, SINDy2} or partial \cite{schaeffer2017learning} differential equation and further extended to tackle stochastic differential equations with Gaussian noise based on Kramers-Moyal expansion \cite{Boninsegna2018}. The Koopman operator theory \cite{EDMD, SKO3} was also employed to develop algorithm to identify dynamical systems from time series data. There are also some data-driven methods based on powerful representative capability of neural network technique to discover governing equations from data, such as neural ordinary differential equations \cite{Duvenaud1, NeuralSDE}, physics-informed neural network \cite{PINNs}, normalizing flows \cite{papamakarios2021normalizing}, to name a few.

These methodologies primarily concentrate on deterministic dynamical systems or those perturbed by Gaussian noise. While non-Gaussian noise has substantial applications, there is a relative scarcity of data-driven methods devoted to discovering non-Gaussian stochastic dynamical systems. Fortunately, our prior research has proposed and proved the nonlocal Kramers-Moyal formulas, which link the L\'evy jump measure, drift coefficient, and diffusion coefficient of non-Gaussian stochastic dynamical systems to their sample paths \cite{YangLi2020a, liyang2021b}. Armed with these formulas, sparse regression methods can be effectively utilized to extract the governing equations of non-Gaussian stochastic dynamical models from sample path data. Nonetheless, selecting an appropriate dictionary of candidate functions for the sparse regression method presents a considerable challenge. Concurrently, normalizing flows have been harnessed to integrate with nonlocal Kramers-Moyal formulas for data-driven modeling \cite{luy2022a, Li_2022}. While the neural network framework offers significant representational power, it faces the limitation of interpretability due to the lack of explicit formulas, which can hinder the understanding and trust in the models it produces.

Striving to harmonize interpretability with the capacity for representation, symbolic regression methods have risen as a suitable selection for data-driven modeling endeavors. Unlike traditional deterministic regression techniques that assume a predefined mathematical structure and merely identify parameters to best fit the data, evolutionary symbolic regression approaches are designed to simultaneously discover parameters and infer the optimal functional form of the model from input-response datasets \cite{udrescu2020ai, angelis2023artificial}. These evolutionary algorithms meticulously seek out functional abstractions, employing predeined mathematical operators and function symbols, and endeavoring to reduce error metrics to their minimum. Genetic programming, prevalent in symbolic regression, is an advanced and generalized version of the genetic algorithm, inspired by Darwin's theory of natural selection \cite{langdon2013foundations, slowik2020evolutionary, north2023review}. The application of genetic programming has been extensively recognized, making significant progress across a spectrum of disciplines, including digital signal processing \cite{yang2005force}, hidden physics detecting \cite{vaddireddy2020feature}, and the estimation of aerodynamic parameters \cite{luo2015adaptive}. Building upon these advancements, Askari and Crevecoeur \cite{askari2023evolutionary} have advanced the field by integrating sparse regression techniques with genetic programming, enabling data-driven modeling of both ordinary and hybrid dynamical systems and expanding the potential for automatically uncovering empirical equations.

Motivated by the work presented in Ref. \cite{askari2023evolutionary}, this article endeavors to devise the ESSR method for discovering non-Gaussian stochastic dynamical systems. Our approach is grounded in nonlocal Kramers-Moyal formulas, genetic programming, and sparse regression, offering a robust framework for data-driven modeling of such systems. The structure of this paper is outlined as follows. In Section \ref{NKMFsec}, we review nonlocal Kramers-Moyal formulas of non-Gaussian stochastic dynamical systems, laying the groundwork for our algorithmic development. In Section \ref{NUMEsec}, we design a data-driven algorithm to effectively discover L\'evy jump measure, drift coefficient and diffusion coefficient from data. The efficacy and precision of our methodology are then demonstrated through a series of numerical experiments in Section \ref{NUEXsec}. We conclude the paper in Section \ref{CONCsec} with a summary of our findings and insights.

\section{Nonlocal Kramers-Moyal formulas}
\label{NKMFsec}
Consider an $n$-dimensional stochastic dynamical system in the following form
\begin{equation} \label{SDE}
	dx\left( t \right)=b\left(x\left( t \right) \right)dt+\sigma_1 \left( x\left( t \right) \right)d{{B}_{t}}+\sigma_2 d{{L}_{t}},
\end{equation}
where ${{B}_{t}}=[B_{1t},B_{2t},...,B_{nt}]^T$  is an $n$-dimensional (Gaussian) Brownian motion,  and ${{L}_{t}}=[L_{1t},L_{2t},...,L_{nt}]^T$ is an $n$-dimensional (non-Gaussian) symmetric L\'evy motion described in the Appendix with the positive constant noise intensity $\sigma_2 $ and the jump measure ${{\nu }}\left( dy \right)={{W}}\left( y \right)dy$, for $y\in \mathbb{R}^{n}\backslash \left\{ 0 \right\}$. The kernel function of the jump measure satisfies $W(-y)=W(y)$ according to the symmetric property. The vector $b\left(x \right)$  is the drift coefficient (or the vector field) in ${{\mathbb{R}}^{n}}$  and the diffusion matrix is $a\left( x \right)=\sigma_1 {{\sigma_1 }^{T}}$. Assume that the initial condition is $x\left( 0 \right)=z$.

As the generalized version of traditional Kramers-Moyal formulas, we proposed and proved the following theorem for the stochastic differential equation (\ref{SDE}) in the previous work \cite{YangLi2020a, liyang2021b}. This theorem aims to express the L\'evy jump measure, drift coefficient and diffusion coefficient in terms of the transition probability density $p\left( x,t|z,0 \right)$ (solution of Fokker-Planck equation).

\newtheorem{thm}{\bf Theorem}
\begin{thm} (Relation between stochastic governing law and Fokker-Planck equation)\\
	\label{thm1}
	For every $\varepsilon >0$, the probability density function $p\left( x,t|z,0 \right)$ and the jump measure, drift and diffusion have the following relations:\\
	1) For every $x$ and $z$ satisfying $\left| x-z \right|>\varepsilon $,
	\begin{align}\label{T.1}
		\underset{t\to 0}{\mathop{\lim }}\,{{t}^{-1}}{p}\left( x,t|z,0 \right)=\sigma_2^{-n}W\left( \sigma_2^{-1}\left( x-z \right) \right)
	\end{align}
	uniformly in $x$ and $z$.\\
	2) For $i=1,\ 2,\ \ldots ,\ n$,
	\begin{align}\label{T.2}
		\underset{t\to 0}{\mathop{\lim }}\,{{t}^{-1}}\int_{\left| x-z \right|<\varepsilon}{\left( {{x}_{i}}-{{z}_{i}} \right)p\left( x,t|z,0 \right) \textrm{d}\mathbf{x}}={{b}_{i}}\left( z \right).
	\end{align}
	3) For $i,j=1,\ 2,\ \ldots ,\ n$,
	\begin{align}\label{T.3}
		\begin{split}
		& \underset{t\to 0}{\mathop{\lim }}\,{{t}^{-1}}\int_{\left| x-z \right|<\varepsilon}{\left( {{x}_{i}}-{{z}_{i}} \right)\left( {{x}_{j}}-{{z}_{j}} \right)p\left( x,t|z,0 \right) \textrm{d}\mathbf{x}}
		\\ & ={{a}_{ij}}\left( z \right) + \sigma_2^{-n}\int_{\left|y\right|<\varepsilon}{{y}_{i}{y}_{j}W\left( {\sigma}_2^{-1}{y} \right) dy}.
		\end{split}
	\end{align}
\end{thm}

For convenience of numerical method, these three formulas can be reformulated as the following corollary, which use the sample path $x(t)$ (solution of the stochastic differential equation (\ref{SDE})) to express the jump measure, drift coefficient and diffusion coefficient. Since they can be regarded as the generalization of traditional Kramers-Moyal formulas, we refer to the corollary as nonlocal Kramers-Moyal formulas \cite{YangLi2020a, liyang2021b}.

\newtheorem{cor}[thm]{\bf Corollary}
\begin{cor}(Nonlocal Kramers-Moyal formulas)\\
	\label{cor2}
	For every $\varepsilon >0$, the sample path solution $x\left( t \right)$ of the stochastic differential equation (\ref{SDE}) and the jump measure, drift and diffusion have the following relations:\\
	1) For every $m>1$,
	\begin{align}\label{C.1}
		\begin{split}
		&\underset{t\to 0}{\mathop{\lim }}\,{{t}^{-1}}\mathbb{P}\left\{ \left. \left| {x}\left( t \right)-{z}\right| \in \left[ \varepsilon,\ m\varepsilon \right) \right| x\left( 0 \right)=z \right\}
		\\ & =\sigma_2 ^{-n} \int_{ \left| y \right| \in \left[ \varepsilon,\ m\varepsilon \right)} {W\left( \sigma_2 ^{-1} {y} \right) dy}.
		\end{split}
	\end{align}
	2) For $i=1,\ 2,\ \ldots ,\ n$,
	\begin{align}\label{C.2}
		\begin{split}
		&\underset{t\to 0}{\mathop{\lim }}\,{{t}^{-1}}\mathbb{P}\left\{ \left. \left| x\left( t \right)-z \right| <\varepsilon  \right| x\left( 0 \right) =z \right\}
		\\ & \cdot \mathbb{E} \left[ \left. \left( {{x}_{i}}\left( t \right) -{{z}_{i}} \right) \right| x\left( 0 \right)=z;\ \left| x\left( t \right)-z \right| <\varepsilon  \right] ={{b}_{i}}\left( z \right).
		\end{split}
	\end{align}
	3) For $i,j=1,\ 2,\ \ldots ,\ n$,
	\begin{align}\label{C.3}
		\begin{split}
		& \underset{t\to 0}{\mathop{\lim }}\,{{t}^{-1}}\mathbb{P}\left\{ \left. \left| x\left( t \right)-z \right| <\varepsilon  \right| x\left( 0 \right) =z \right\} \\
		& \cdot \mathbb{E} \left[ \left. \left( {{x}_{i}}\left( t \right) -{{z}_{i}} \right) \left( {{x}_{j}}\left( t \right) -{{z}_{j}} \right) \right| x\left( 0 \right)=z;\ \left| x\left( t \right)-z \right| <\varepsilon  \right] \\
		& ={{a}_{ij}}\left( z \right)+\sigma_2^{-n}\int_{\left|y\right|<\varepsilon}{{y}_{i}{y}_{j}W\left( {\sigma}_2^{-1}{y} \right) dy}. 
		\end{split}
	\end{align}
\end{cor}

Based on the nonlocal Kramers-Moyal formulas and sparse learning method, we designed numerical schemes to discover the L\'evy jump measure, drift coefficient and diffusion coefficient of stochastic differential equations from sample path data in the previous work \cite{YangLi2020a, liyang2021b}. However, the sparse learning method requires proper choice of candidate functions, which relies on the prior knowledge of the underlying system. Thus we construct our algorithm by introducing genetic programming technique to avoid this problem in the next section.

\section{Numerical method}
\label{NUMEsec}
\begin{figure}
	\centering
	\includegraphics[width=7cm]{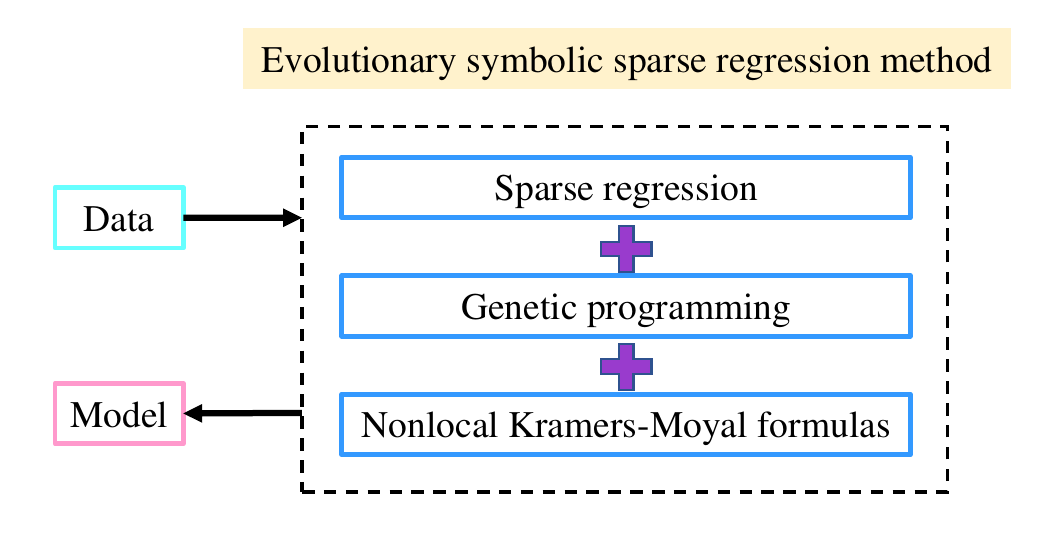}
	\caption{Schematic diagram of the basic structure of the ESSR method.}
	\label{fig1}
\end{figure}

This research aims at developing the ESSR method to discover stochastic dynamical systems with both (Gaussian) Brown noise and (non-Gaussian) L\'evy noise from sample path data without prior knowledge of system information. As shown in Fig. \ref{fig1}, the core idea of the evolutionary symbolic sparse regression method is the combination of sparse regression, genetic programming and nonlocal Kramers-Moyal formulas. Roughly speaking, the role of genetic programming algorithm is to generate candidate functions, the sparse regression technique serves as learning the coefficients of these candidate functions, and the nonlocal Kramers-Moyal formulas are used to construct the fitness function in genetic programming and the loss function in sparse regression. The input of the method is sample path data of the system variable and its output is learned model of non-Gaussian stochastic differential equation (\ref{SDE}), i.e., drift coefficient $b(x)$, diffusion coefficient $a(x)$, kernel function $W(x)$ of L\'evy jump measure and constant L\'evy noise intensity $\sigma_2$. For convenience of numerical computation, we assume that the L\'evy process $L_t$ is rotationally symmetric, $W(x)=W(|x|)$. This section will introduce framework and details of this method.

\subsection{Data preprocessing}
\label{DPPsec}
Assume that the phase space of interest is $\mathrm{\Omega}$ and we have a pair of datasets $Z=\{z_1,z_2,...,z_M\}$ and $X=\{x_1,x_2,...,x_M\}$ with $M$ elements, where $z_i$ and $x_i$ are the observational data of the state variable $x(t)$ (solution of Eq. (\ref{SDE})) at initial time $t=0$ and a small time $t=h$.  In order to construct the subsequent fitness function and loss function in our algorithm, we firstly need to perform data preprocessing procedure to generate the training datasets $\mathbb{X}$ and $\mathbb{Y}$ based on the sample path datasets $Z$ and $X$.

First of all, we begin by constructing the training datasets $\mathbb{X}_1$ and $\mathbb{Y}_1$ to learn L\'evy jump measure and noise intensity, utilizing the first equation (\ref{C.1}) of nonlocal Kramers-Moyal formulas. A keen observation from Eq. (\ref{C.1}) reveals that the the dependency of both sides of the equation is solely on the norm of the difference between $z$ and $x$, rather than their specific positions. To leverage this insight, we construct a dataset $R=\{ |y_1|,|y_2|,...,|y_M| \}$ where $y_i=x_i-z_i$ for $i=1,2,...,M$. The conditional probability of left side of Eq. (\ref{C.1}) can be estimated as the proportion of points in the set $R$ that fall within the interval $\left[ \varepsilon, m\varepsilon \right)$ relative to the total count $M$. Considering $N$ intervals deined as $\left[ \varepsilon, m\varepsilon \right)$, $\left[ m\varepsilon, m^2\varepsilon \right)$,..., $\left[ m^{N-1}\varepsilon, m^N\varepsilon \right)$ and assuming that the number of points falling into these intervals are $n_1$, $n_2$,..., $n_N$, respectively, the left-hand side of Eq. (\ref{C.1}) can be approximated as
\begin{equation*}
	{{h}^{-1}}\mathbb{P}\left\{ \left. \left| {x}\left( t \right)-{z}\right| \in \left[ m^{i-1}\varepsilon,\ m^i\varepsilon \right) \right| x\left( 0 \right)=z \right\} \approx h^{-1}M^{-1}n_i, \ i=1,2,...,N.
\end{equation*}
Therefore, we define the training datasets as $\mathbb{X}_1 = \{ \varepsilon, m\varepsilon,..., m^N\varepsilon \}$ with $N+1$ elements and $\mathbb{Y}_1 = \{ h^{-1}M^{-1}n_1, h^{-1}M^{-1}n_2,..., h^{-1}M^{-1}n_N \}$ with $N$ elements.

Secondly, we proceed to construct the training datasets $\mathbb{X}_2$ and $\mathbb{Y}_2$ for the purpose of extracting drift coefficient, which is derived from the second equation (\ref{C.2}) of nonlocal Kramers-Moyal formulas. Given that the subsequent genetic programming technique is highly sensitive to noise, we divide the phase space $\mathrm{\Omega}$ into $N_b$ smaller bins, denoted as $\mathrm{\Omega}_k$, $k=1,2,...,N_b$. In accordance with this discretization, the datasets $Z$ and $X$ are correspondingly segmented into datasets $Z_k$ and $X_k$ for each $k=1,2,...,N_b$. Within the confines of each bin $\mathrm{\Omega}_k$, the conditional probability in Eq. (\ref{C.2}) is estimated by calculating the ratio of the number of points satisfying the condition $\left| {x}\left( t \right)-{z}\right| \leq \varepsilon$ to the total count of points in $Z_k$, which we designate as $p_k$. Assuming that the datasets $\hat{Z}_k$ and $\hat{X}_k$ are obtained by excluding the data from $Z_k$ and $X_k$ satisfying $\left| {x}\left( t \right)-{z}\right| > \varepsilon$, we then employ a linear approximation coupled with the least squares method to calculate the drift value $b(z_k)$ at the central point $z_k$ of the bin $\mathrm{\Omega}_k$ utilizing the datasets $\hat{Z}_k$ and $\hat{X}_k$. To elaborate, we assume that the stochastic system possesses ergodic property and consider the dimension $i$, $i=1,2,...,n$. Then we have
\begin{equation}\label{eqbin1}
	\begin{aligned}
	& Ac_i=B_i, \\
	& A= \begin{bmatrix}
		1 & \hat{z}_{1,1} & \cdots & \hat{z}_{n,1} \\
		\vdots & \vdots & \ddots & \vdots \\
		1 & \hat{z}_{1,\hat{M}_k} & \cdots & \hat{z}_{n,\hat{M}_k}
	\end{bmatrix}, \\
	& B_i= p_k h^{-1} \left[ x_{i,1}-z_{i,1},..., x_{i,\hat{M}_k}-z_{i,\hat{M}_k} \right]^T,
	\end{aligned}
\end{equation}
where $\hat{M}_k$ is the number of elements in $\hat{Z}_k$. Note that the optimal solution of Eq. (\ref{eqbin1}) in the sense of least squares is $\tilde{c}_i=(A^TA)^{-1}A^TB_i$. Thus we obtain a pair of training data within the bin $\mathrm{\Omega}_k$ by linear approximation, i.e., the central point  $z_k$ and its drift value $b_{i,k}=(1, z_{1,k},..., z_{n,k})\tilde{c}_i$. Therefore, we define the training datasets for drift coefficient as $\mathbb{X}_2 = \{ z_1, z_2,..., z_{Nb} \}$ and $\mathbb{Y}_2 = \{ b_1, b_2,..., b_{Nb} \}$.

Thirdly, we advance to the construction of the training datasets $\mathbb{X}_3$ and $\mathbb{Y}_3$, which are pivotal for extracting the diffusion coefficient, as dictated by the third equation (\ref{C.3}) of nonlocal Kramers-Moyal formulas. This process mirrors the previous operation concerning the drift term, with a key distinction: it necessitates the substitution of the identified L\'evy jump measures and noise intensity into the right-hand side of equation (\ref{C.3}). For the sake of clarity and convenience, we denote this integral as $S_{ij}(\varepsilon)$. Utilizing the discretization previously described, we obtain the following equation specific to each bin $\mathrm{\Omega}_k$, applicable for $i,j=1,2,...,n$
\begin{equation}\label{eqbin2}
	\begin{aligned}
		& Ad_{ij}=B_{ij}, \\
		& A= \begin{bmatrix}
			1 & \hat{z}_{1,1} & \cdots & \hat{z}_{n,1} \\
			\vdots & \vdots & \ddots & \vdots \\
			1 & \hat{z}_{1,\hat{M}_k} & \cdots & \hat{z}_{n,\hat{M}_k}
		\end{bmatrix}, \\
		& B_{ij}= p_k h^{-1} \left[ (x_{i,1}-z_{i,1})(x_{j,1}-z_{j,1}),..., (x_{i,\hat{M}_k}-z_{i,\hat{M}_k})(x_{j,\hat{M}_k}-z_{j,\hat{M}_k}) \right]^T -S_{ij}.
	\end{aligned}
\end{equation}
Similarly, the optimal solution of Eq. (\ref{eqbin2}) in the context of least squares is $\tilde{d}_{ij}=(A^TA)^{-1}A^TB_{ij}$. Consequently, a pair of training data within each bin $\mathrm{\Omega}_k$ is determined thorugh a linear approximation. Specifically, this involves the central point  $z_k$ and its associated diffusion value, calculated as $a_{ij,k}=(1, z_{1,k},..., z_{n,k})\tilde{d}_{ij}$. Therefore, we define the training datasets for diffusion coefficient with the following structures: $\mathbb{X}_3 = \{ z_1, z_2,..., z_{Nb} \}$ and $\mathbb{Y}_3 = \{ a_1, a_2,..., a_{Nb} \}$.

\subsection{Sparse regression}
\label{SRsec}
Let us assume that we possess a set of $N_c$ candidate functions, denoted as $\phi_1(x)$, $\phi_2(x)$,..., $\phi_{N_c}(x)$. The linear combinations of these functions, represented as $[\phi_1, \phi_2, ..., \phi_{N_c}] \xi$, serve as the foundation for approximating the L\'evy jump measure, drift coefficient and diffusion coefficient. These candidate functions are to be generated by a genetic programming algorithm, the details of which will be elucidated subsequently. The undetermined coefficient vector $\xi$ plays a pivotal role in determining the activation of the candidate functions and the degree to which they approximate the unknown function. 

To begin with, we delve into the sparse regression process for the L\'evy jump measure. Given that L\'evy motion is presumed to exhibit rotational symmetry, the right-hand side of Eq. (\ref{C.1}) can be re-expressed through a polar coordinate transformation for various dimensions, yielding the following integrals
\begin{equation}\label{jm123}
	\begin{aligned}
	& \int_{ r \in \left[ \varepsilon,\ m\varepsilon \right)} 2\sigma_2 ^{-n} {W\left( \sigma_2 ^{-1} {r} \right) dr}, n=1, \\
	& \int_{ r \in \left[ \varepsilon,\ m\varepsilon \right)} 2 \pi \sigma_2 ^{-n} r{W\left( \sigma_2 ^{-1} {r} \right) dr}, n=2, \\
	& \int_{ r \in \left[ \varepsilon,\ m\varepsilon \right)} 4 \pi \sigma_2 ^{-n} r^2 {W\left( \sigma_2 ^{-1} {r} \right) dr}, n=3, \\
	& ...
	\end{aligned}
\end{equation}
Assuming that the integrand within these integrals can be approximated by a linear combination of candidate functions $[\phi_1, \phi_2, ..., \phi_{N_c}] \xi_1$, we can then formulate a series of equations in accordance with Eq. (\ref{C.1})
\begin{equation}\label{linear1}
	\begin{aligned}
		& \left[\int_{ r \in \left[ m^{i-1}\varepsilon,\ m^i\varepsilon \right)} \phi_1(r) dr, ..., \int_{ r \in \left[ m^{i-1}\varepsilon,\ m^i\varepsilon \right)} \phi_{N_c}(r) dr \right] \xi_1 \\
		& = h^{-1} M^{-1} n_i, \ \ \ i=1,2,...,N. 
	\end{aligned}
\end{equation}
Denote this group of equations as $\mathrm{\Phi}_{\mathbb{X}_1}\xi_1 =\mathbb{Y}_1$ and then we define the error function as
\begin{equation}\label{error1}
	\mathcal{L}_1 = \frac{1}{N} |\mathrm{\Phi}_{\mathbb{X}_1}\xi_1-\mathbb{Y}_1|^2.
\end{equation}

Furthermore, by approximating the drift function as a linear combination of candidate functions, $[\phi_1, \phi_2, ..., \phi_{N_c}] \xi_2$, we can obtain the following equations utilizing the training datasets $\mathbb{X}_2$ and $\mathbb{Y}_2$
\begin{equation}\label{linear2}
	\begin{aligned}
		\left[\phi_1(z_k), ..., \phi_{N_c}(z_k) \right] \xi_2 =b_{i,k}, \ k=1,2,...,N_b, \ i=1,2,...,n. 
	\end{aligned}
\end{equation}
Denote this group of equations as $\mathrm{\Phi}_{\mathbb{X}_2}\xi_2 =\mathbb{Y}_2$ and then we can define the error function for drift term as
\begin{equation}\label{error2}
	\mathcal{L}_2 = \frac{1}{N_b} |\mathrm{\Phi}_{\mathbb{X}_2}\xi_2-\mathbb{Y}_2|^2.
\end{equation}
This error function is designed to encapsulate the discrepancies in the approximation of the drift term, allowing for the assessment of the model's performance across either a single dimension or a combination of several dimensions.

Following the same methodology, let us assume that the diffusion function is represented by a linear combination of the form $[\phi_1, \phi_2, ..., \phi_{N_c}] \xi_3$, and then we can establish the following equations using the training datasets $\mathbb{X}_3$ and $\mathbb{Y}_3$
\begin{equation}\label{linear3}
	\begin{aligned}
		\left[\phi_1(z_k), ..., \phi_{N_c}(z_k) \right] \xi_3 =a_{ij,k}, \ k=1,2,...,N_b, \ i,j=1,2,...,n. 
	\end{aligned}
\end{equation}
We denote this series of equations as $\mathrm{\Phi}_{\mathbb{X}_3}\xi_3 =\mathbb{Y}_3$ and then we define the error function for diffusion coefficient as
\begin{equation}\label{error3}
	\mathcal{L}_3 = \frac{1}{N_b} |\mathrm{\Phi}_{\mathbb{X}_3}\xi_3-\mathbb{Y}_3|^2.
\end{equation}

The minimization of the objective functions (\ref{error1}), (\ref{error2}) and (\ref{error3}) is pivotal for obtaining the optimal solutions of jump measure, drift term and diffusion term, respectively. However, in the general least squares solution, given by $\xi_i= (\mathrm{\Phi}_{\mathbb{X}_i}^T \mathrm{\Phi}_{\mathbb{X}_i})^{-1} \mathrm{\Phi}_{\mathbb{X}_i}^T \mathbb{Y}_i$, each component tends to be non-zero, which can complicate subsequent analysis and interpretation of the model. Moreover, it contradicts the criterion that the discovered model should be parsimonious as far as possible. To address these challenges, researchers have formulated various regularizing regression techniques that impose penalties on non-zero coefficients, thereby promoting a more parsimonious model, including sequential threshold ridge regression \cite{saleh2019theory}, the least absolute shrinkage and selection operator (lasso) \cite{kukreja2006least}, and the elastic-net approach \cite{de2009elastic}.

In this study, we employ the elastic-net method to foster sparsity within our model, which is encapsulated by the following loss function
\begin{equation}\label{loss}
	T_{\lambda} = \mathcal{L} (\xi) +\lambda J(\xi).
\end{equation}
Here $\lambda$ acts as the regularization parameter, dictating the degree of penalization applied to the coefficients. The penalty function $J(\xi)$ is defined as $J(\xi)= \beta |\xi|_2^2 + (1-\beta) |\xi|_1$, where $\beta$ is a mixing parameter that allows for a blend between two types of penalties: the $L_2$ norm, which is associated with ridge regression when $\beta=1$, and the $L_1$ norm, which corresponds to lasso regression when $\beta=0$. For the sake of simplicity, we have omitted the subscript $i$ from $\mathcal{L}$ in this context. The elastic-net approach provides a flexible framework that combines the best features of both ridge and lasso regressions, offering a balance between the bias-variance trade-off and the ability to select relevant features. In practical terms, the elastic-net method can be conveniently implemented using the \textit{lasso} function available in Matlab, which allows for the optimization of the loss function and the determination of the optimal values for $\xi$ that minimize the trade-off between model complexity and fitting accuracy.

\subsection{Genetic programming}
\label{GPsec}

Symbolic regression is the process of uncovering a mathematical expression in symbolic form that optimally fits the provided datasets. This methodology stands in contrast to traditional regression techniques, where the model's structure is pre-defined and the focus is on estimating the model's unknown parameters. Genetic programming is a prominent approach within symbolic regression, applicable to a variety of domains such as optimal control, data-driven modeling, automatic programming, classification, and engineering design. It extends the binary representation used in genetic algorithms to a more complex tree-based structure \cite{koza1994genetic}. Substantially, genetic programming follows Darwin's evolution theory, selecting the individle that fits the environment (observational data) the best via evolution of generation (iterating algorithmic process) and natural selection. In this paper, the genetic programming aims to generate a set of candidate functions which is provided for sparse regression to constitute the physics-based model. For a more comprehensive overview of genetic programming, see the references \cite{koza1994genetic}.

\begin{figure}
	\centering
	\includegraphics[width=7cm]{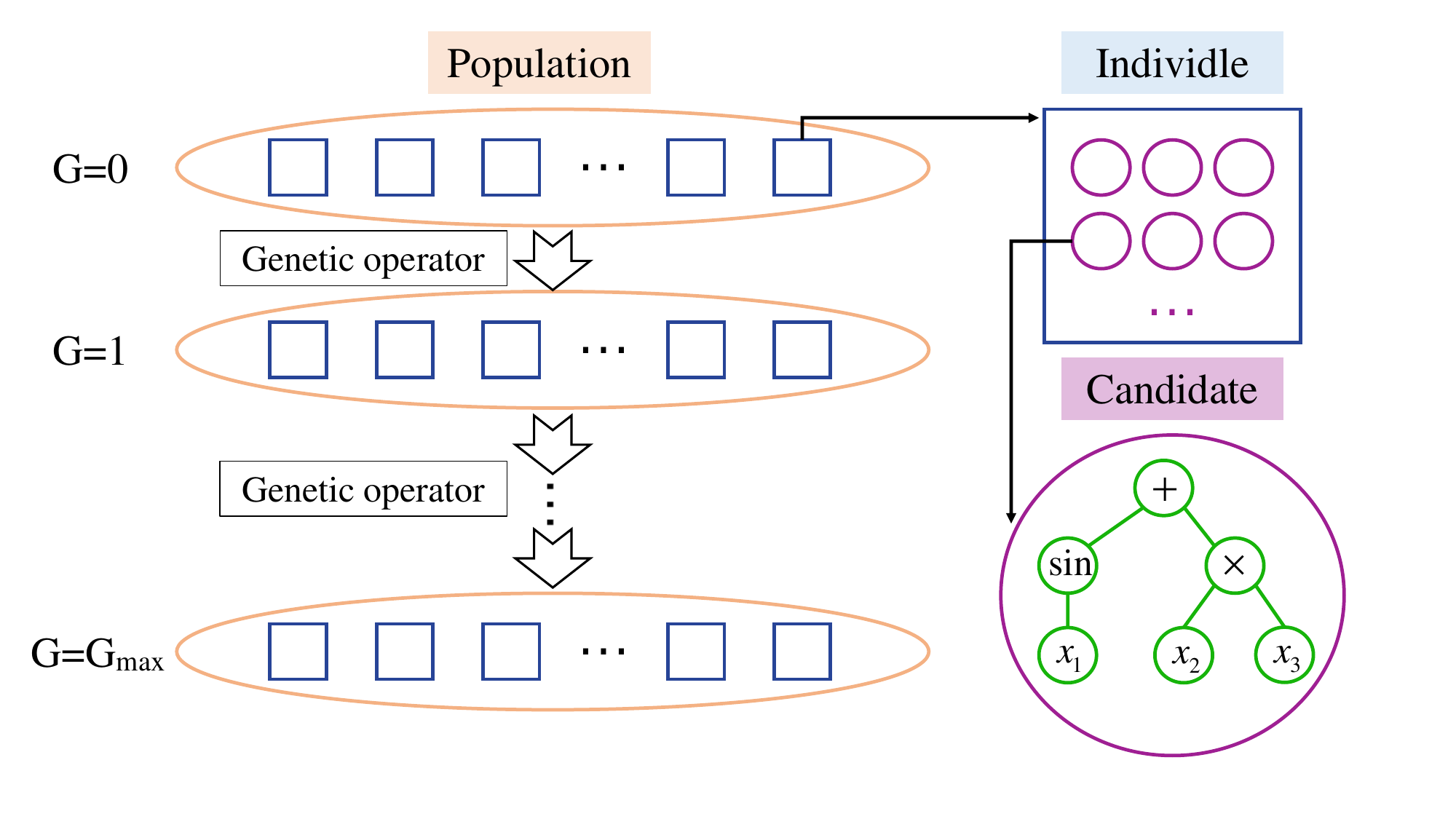}
	\caption{Framework of genetic programming. Gold ellipse denotes population of genetic programming and $G$ indicates the generation. Blue box represents individle within the population, which is constituted by several candidate functions indicated by purple circle. The candidate functions are mathematical expressions, composed of the operators in the function set and variables in the terminal set.}
	\label{fig2}
\end{figure}

As depicted in Figure \ref{fig2}, the genetic programming framework encompasses several key stages: the generation of an initial random population, the iterative selection of individuals based on a predefined fitness function, and the improvement of individuals through the application of one or more genetic operators. It is assumed that the population is composed of $N_P$ individles, and the algorithm is designed to iterate for a maximum of $G_{{max}}$ generations. The process continues until convergence is achieved or the maximum number of generations is reached. Diverging from the conventional genetic programming paradigm, where each individual encompasses a single candidate function, this paper posits that an individual may contain multiple candidate functions, inspired by prior research \cite{askari2023evolutionary}. Consequently, the linear combination of these candidates within a single individual is utilized to approximate the physics-based model.

Each candidate function is a mathematical expression, created by a tree structure. It is composed of the operators in the function set and variables in the terminal set. In this research, we define the two sets as the following structures
\begin{equation}\label{functionset}
	\{ +, \times, /, \exp, \ln, \sin, \text{squ}, ... \},
\end{equation}
\begin{equation}\label{terminalset}
	\{ 1, c, x_1, x_2, ..., x_n \}.
\end{equation}
The terminal set (\ref{terminalset}) contains the numeral 1, the constant $c$, and all the state variables $x_i$, $i=1,2,...,n$. The function set, on the other hand,  (\ref{functionset}) comprises mathematical operations and a selection of elementary function symbols. The exclusion of subtraction from the function set is deliberate, as it can be alternately represented as "$+(-1)$", with -1 being expressible through the constant $c$ in the terminal set. The symbol "$\text{squ}$" in the function set denotes the square function $\text{squ}(x)=x^2$, which can simplify the presentation of high-order polynomial functions. For example, $x^4$ can be represented as $\text{squ}(\text{squ}(x))$ with 3 nodes while it requires 7 nodes for $x \times x \times x \times x$. It is important to recognize that a wide array of common functions can be synthesized through the composition of elements within the function set. For instance, $|x| ^ \alpha= \exp \{ 0.5 \times \alpha \ln [\text{squ}(x)] \}$ for $\alpha \in \mathbb{R} \backslash 0$ and $\cos{\omega x}= \sin(\omega \times x+ \frac{\pi}{2})$. In the following, we will delve into the foundational concepts and procedural aspects of genetic programming. This will include discussions on the generation of the initial population, the application of genetic operators, and the assessment of fitness within the genetic programming paradigm.

\subsubsection{Initial population generation}
\label{IPGsec}
As illustrated in the bottom-right section of Figure \ref{fig2}, the candidate functions are mathematical expressions that are instantiated by tree structures. In these trees, the terminal nodes are selected from the terminal set, while the non-terminal nodes are derived from the function set. The mathematical operations, such as addition ($+$), are binary operators that have two subnodes, whereas elementary function symbols, such as sine ($\sin$), are unary operators that possess a single subnode. It is important to highlight that each constant $c$ that appears within the tree structure is initially assigned a random value, confined within a user-defined interval $[c_{min},c_{max}]$. Throughout the iterative process of genetic programming, the values assigned to these constants remain constant unless the individual constants are selected to participate in crossover or mutation operations.

The initial population in genetic programming (the 0th generation) consists of $N_P$ individuals that are randomly generated. The act of generating this initial population can be conceptualized as a form of blind search across the program space. When assembling the initial individuals, function operators and terminal symbols are indiscriminately selected from the respective function and terminal sets. If a function operator is selected, the process is recursively repeated for its sub-nodes; if a terminal symbol is selected, the development of that particular subtree is concluded. This recursive procedure persists until each branch in the tree culminates in a terminal symbol as its sub-node. There are two principal methodologies for generating the initial population: one involves fixing the depth or size of the nodes for the candidate functions, while the other involves random generation of the candidate functions with only a maximum depth or maximum node count specified for the initial population. In this paper, we opt for the former approach, ensuring that each initial candidate function is structured with exactly 5 nodes.

\subsubsection{Genetic operators}
\label{GOsec}
As illustrated in Fig. \ref{fig2}, once the initial population has been established, it must be subjected to iterative enhancement through the application of genetic operators to the individuals within the population. These genetic operators can be categorized into five types: reproduction, crossover, mutation for subtree, mutation for constant, editing. We will give a detailed description of them used in genetic programming in the following.

According to Darwin's evolution theory, individuals that are better adapted to their environment have a higher probability of leaving offspring. Therefore, a portion of the individuals with better fitness should undergo a \textbf{reproduction} operator, directly copying them into the next generation. We define a new set $P_b^G$, called the set of the best individles, to deposit these reproduction individles. The reproduction operation is carried out via two schemes. The former is to directly reproduce the individles with the lowest fitness for a proportion of $\varepsilon_1 \%$ of the current population, commonly set at $2\%$. The second scheme, known as tournament selection, randomly and uniformly selects a small group of individuals (usually two or three) from the population, excluding the members of $P_b^G$, and reproduces the individle with the lowest fitness within this group. The percentage of the later selection method is $\varepsilon_2 \%$, commonly $18\%$. After the set $P_b^G$ is constructed, it is reproduced to the next generation. That is, the new generation contains $(\varepsilon_1+\varepsilon_2)\%$ of the previous generation.

\begin{figure}
	\centering
	\includegraphics[width=11cm]{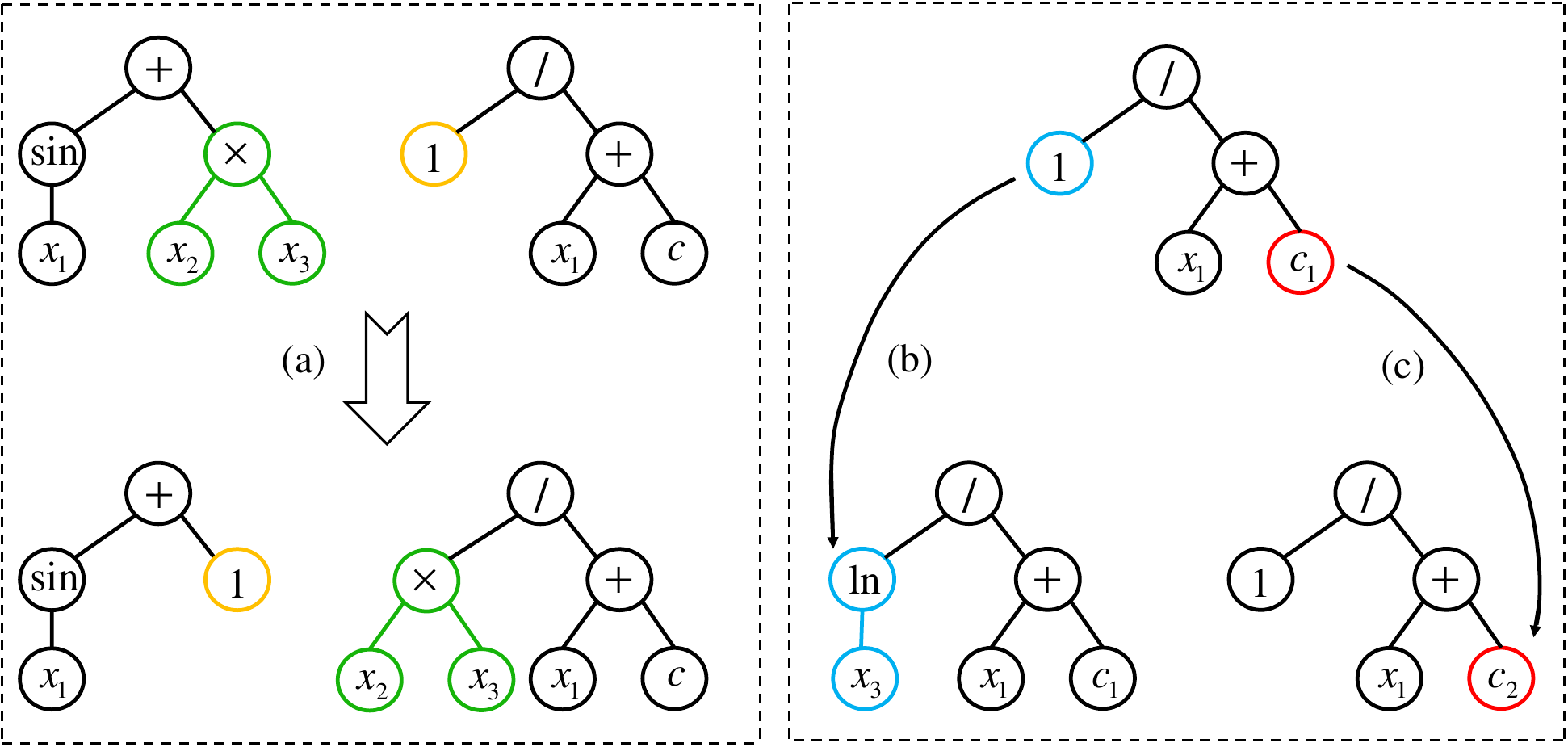}
	\caption{Genetic operators of genetic programming. (a) Crossover. Two parental candidate functions, $\sin x_1 + x_2 x_3$ and $1/(x_1+c)$, produce two offspring trees $\sin x_1 + 1$ and $x_2 x_3/(x_1+c)$ after crossover operation. (b) Mutation for subtree. One parental expression $1/(x_1+c_1)$ produces an offspring tree $\ln x_1 /(x_1+c_1)$ after mutation for subtree. (c) Mutation for constant. One parental expression $1/(x_1+c_1)$ produces an offspring tree $1 /(x_1+c_2)$ after mutation for constant.}
	\label{fig3}
\end{figure}

The \textbf{crossover} operator is a principal mechanism for generating new individles in the population, which randomly chooses two parents and produces two offsprings through the exchange of certain parts, as demonstrated in Fig. \ref{fig3}(a). Specifically, we employ the tournament selection method to choose two individles from $P_b^G$,respectively. It is important to note that within each individual, two functions are randomly chosen to serve as the parental expressions, provided that they are not simultaneously operands (elements of the terminal set). The selection process for candidate functions within individuals and for nodes within these candidates follows a uniform probability distribution. The choice of crossover fragments can range from entire trees to just a single operand. Once the crossover points have been identified, we proceed to exchange the two crossover nodes, including the entire subtree beneath them. Except for the two parental expressions, other candidate functions in the two individles remain unchanged entering the offspring individles.

The operation of \textbf{mutation for subtree} is specifically designed to introduce new genetic variations and enhance the diversity within the population structure, thereby preventing the algorithm converging prematurely. This process begins by randomly selecting a parental individual from the set of best individuals $P_b^G$, using the tournament selection method, which then serves as the basis for generating a single offspring tree. Once the parental individual is selected, a mutation node is randomly chosen from within a randomly selected candidate function of the individual, following a uniform distribution. The entire subtree rooted at the mutation node is then replaced with a new, randomly generated subtree, as depicted in Fig. \ref{fig3}(b). Except for the chosen parental function, other candidates in the individle directly reproduced into the new offspring.

The operation  of \textbf{mutation for constant} is another mutation strategy that aims to fine-tune the constant values within the mathematical expressions, guiding them closer to their true values. It is observed from Fig. \ref{fig3}(c) that this operation targets a constant, say $c_1$, within a randomly selected parental expression and alters it to a new value, $c_2$. Given that this operation is applied to the best individual set, even a minor adjustment can potentially enhance the performance of the identified function terms. Specifically, the mutation for constant is carried out via modifying the constant value according to the following formula \cite{askari2023evolutionary}
\begin{equation}\label{mutationconstant}
	c_{new}=c_{old}+\frac{2\kappa - \vartheta}{\vartheta + G}, \ \kappa \in \left[0,\vartheta \right),
\end{equation}
where $G$ denotes the generation number and $\kappa$ is an random value between $0$ and $\vartheta$ set by the user.

After the successful implementation of the aforementioned genetic operators, we arrive at a new population comprising all $N_P$ individles. Nonetheless, the generated individuals may require further refinement through a process known as the operation of \textbf{editing}, if the following situations are fulfilled:
\begin{itemize}
	\item [1)] Node size reduction: Remove the candidate functions with size of node more than 15.
	\item [2)] Term simplification: We merge the expression $\sin\sin$ as $\sin$ and merge $\exp\exp\exp$, $\ln\ln\ln$ as $\exp\exp$, $\ln\ln$, respectively.
	\item [3)] Independence assurance: Remove the candidate functions within one individle which are not independent, including repeated functions, since dependent functions will lead to ill-conditioned computation of sparse regression.
	\item [4)] Logarithmic argument check: Remove the candidate functions when they have a negetive value inside the logarithmic function $\ln(\cdot)$.
	\item [5)] Rational function verification: Remove the rational candidate functions when a zero denominator emerges.
	\item [6)] Magnitude constraint: Remove the candidate functions if their magnitudes exceed $10^{50}$, to avoid excessively large numbers that may cause numerical instability or overflow.
\end{itemize}

\subsubsection{Fitness function}
\label{FITsec}
In accordance with Darwin's theory of evolution, the trajectory of evolution is primarily governed by the natural selection exerted by the environment. The fitness function serves as the intrinsic catalyst for natural selection, a concept that applies equally to biological systems and to genetic programming. A lower fitness score signifies a better adaptation to the environment or a more accurate alignment with the observed data, thereby increasing the likelihood of that individual producing offspring. This, in turn, shapes the population structure and influences the application of genetic operators within genetic programming.

Our goal is to find the individual that best approximates the training dataset. Formally, the mean squared loss function $\mathcal{L}_I (\xi) = \frac{1}{N} |\mathrm{\Phi}^I_{\mathbb{X}}\xi -\mathbb{Y}|^2$, as detailed in section \ref{SRsec}, quantifies the discrepancy between the output data and the approximation derived from the basis functions. These functions are drawn from the candidate functions within the $I$-th individle, and the coefficient $\xi$ is determined through sparse regression, as outlined in (\ref{loss}). Moreover, it is advantageous for the model we discover to be concise and sparse, as overly complex models can complicate subsequent analysis and comprehension. To this end, regularization terms that account for the number of candidate functions and the number of nodes should be integrated into the fitness function to balance the model's accuracy and complexity.

An adaptive fitness function for the $I$-th individle is defined as follows \cite{askari2023evolutionary}
\begin{equation}\label{fitness}
	\mathcal{F}_I = \frac{\mathcal{L}_I}{\max {\mathcal{L}_b}} \left[ \left( \frac{\Gamma_I}{\max {\Gamma_b}} \right)^{\tau_1} \left( \frac{\Sigma_J \Lambda_{I,J}}{\max {\Lambda_b}} \right)^{\tau_2} \right].
\end{equation}
In this formulation, the subscript "$b$" indicates the set of the best individles, $P_b^G$, and all maximizations are conducted within this set. The variable $\Gamma_I$ records the number of active candidate functions with nonzero coefficients within $I$-th individle and it is normalized by $\max {\Gamma_b}$. Additionally, $\Lambda_{I,J}$ signifies the count of operators within the $J$-th candidate of the $I$-th individle, with the summation extending over all candidates within the $I$-th individle. The operator number is also normalized by the maximal values in the set $P_b^G$. These two terms along with their power exponents $\tau_1$ and $\tau_2$ control the number and size of discovered candidate functions, respectively.

In order to obtain appropriate value of the hyperparameter $\tau_1$, we introduce two threshold variables $N_{thre}$ and $E_{thre}$. If the generation of the algorithm is smaller than $N_{thre}$, then we set $\tau_1=0$ to allow the initial exploration of the solution space without penalizing candidate complexity. When the generation exceeds $N_{thre}$ and the loss function $\mathcal{L}$ of the best individle is greater than $E_{thre}$, then we set $\tau_1=1$ to begin penalizing complexity. Once the minimal loss achieved falls below $E_{thre}$, we increment $\tau_1$ by a small positive value $\delta \tau$, i.e., $\tau^{G+1}_1=\tau^G_1+\delta \tau$, to gradually increase the pressure for simplicity. If the mean square loss experiences a sudden increase, such as surpassing 1.2 times the loss of the previous generation, we retract $\tau_1$ by a larger step,  $\tau^{G+1}_1= \tau^G_1- 3\delta \tau$ and let $\delta \tau :=\delta \tau/2$. Should the loss remain higher than the previous optimal value, we further decrease $\tau_1$ in the subsequent generation, i.e., $\tau^{G+2}_1=\tau^{G+1}_1- 20\delta \tau$.

\subsection{Hard-thresholding ridge regression}
\label{HTRRsec}
In addition to employing sparse regression and incorporating a regularization term for the number of candidates in the fitness measure to encourage parsimony, it is also necessary to utilize hard-thresholding ridge regression to eliminate redundant candidates within the set of best individuals $P_b^G$. To do this, we define a small threshold $\rho$ such as $\rho=0.0001$ and proceed to remove the candidate functions that meet the following condition
\begin{equation}\label{htrr}
	\{ \phi_J \in (\phi_1, \phi_2, ..., \phi_{N_c}) | \xi_J \leq \rho \xi_{max} \},
\end{equation}
where $\xi_J$ represents the coefficient, obtained via the least squares method, of the function $\phi_J$ which belongs to one individle and $\xi_{max}=\max_J {\xi_J}$. If there are more than one function satisfying this condition for removal, then the algorithm performs an one-by-one removal scenario.

\subsection{Algorithm}
\label{ALOsec}
In this subsection, we summarize the previous described techniques and procedures into a complete algorithm, as shown in Table \ref{algorithm}. The algorithm takes as input the datasets $Z$, $X$, and all the hyperparameters that have been previously defined. It yields the best individual and its sparse coefficients upon meeting a sufficiently small mean squared error threshold $e_{thre}$ or upon reaching the maximum generation $G_{max}$. To thoroughly discover the information of the system, We run the algorithm three times to learn the L\'evy jump measure, drift coefficient and diffusion coefficient, respectively.

\begin{table*}[htbp]
	\centering
	\caption{ESSR algorithm for learning non-Gaussian stochastic dynamical systems}
	\begin{tabular}{l}
		\hline
		\textbf{Input}: Datasets $Z$ and $X$, all the hyperparameters. \\
		\textbf{Output}: Best individle and its sparse coefficients. \\
		\hline
		
		1. \ Data preprocessing. Construct training datasets $\mathbb{X}_i$ and $\mathbb{Y}_i$, i=1,2,3. \\
		
		2. \ Create initial random population ($G=0$) with $N_P$ individles. \\
		
		3. \ Sparse regression to determine coefficients for all individles. \\
		
		4. \ Evaluate mean squared loss and fitness functions for all individles. \\
		
		5. \ Determine if termination criterions are satisfied, i.e., sufficiently small error or reaching $G_{max}$ generation. \\
		
		6. \ If Yes, then End; If No, then perform reproduction operator. \\
		
		7. \ Construct the best individle set $P_b^G$. \\
		
		8. \ Perform hard-thresholding ridge regression. \\
		
		9. \ The initial index $i= (\varepsilon_1+ \varepsilon_2) \% \times N_P +1$.\\
		
		10. Select genetic operators according to their ratios. \\
		
		11. If crossover, select two individles from $P_b^G$ and generate two new individles; $i=i+2$. \\
		
		12. If mutation for subtree, select one individle from $P_b^G$ and generate one new individle; $i=i+1$. \\
		
		13. If mutation for constant, select one individle from $P_b^G$ and generate one new individle; $i=i+1$. \\
		
		14. Perform editing for new individles. \\
		
		15. If $i<N_P$, back to Step 10; If $i=N_P$, then $G=G+1$ and back to Step 3. \\
		
		\hline
	\end{tabular}
	\label{algorithm}
\end{table*}

\subsection{Hyperparameter selection}
\label{HSsec}
The proposed algorithm is governed by a multitude of hyperparameters that must be predefined. In this subsection, we describe the criteria for selecting these hyperparameters.

In subsection \ref{DPPsec}, we fix $\varepsilon=1$, $m=1.5$, $N=10$ such that there are 10 pairs of data available for estimating the jump measure. The choice of $N_b$ depends on the dimension of system; it typically ranges from dozens for a single dimension to hundreds for two or three dimensions.

Moving on to subsection \ref{SRsec}, The parameters $\lambda$ and $\beta$ are initially set to $\lambda=0.001$ and $\beta=0.8$ during the iterative phase of the algorithm. If the algorithm finds the optimal solution and the model sparsity is not sufficient, then these parameters are adjusted to further enhance the model's parsimony.

In the hard-thresholding ridge regression subsection, we fix the threshold $\rho$ as $\rho=0.0001$ in this paper.

In the genetic programming part, the parameters $G_{max}$, $N_P$ and $e_{thre}$ are selected according to complexity and dimension of the system. The range of values for the constant is set to $[-10,10]$, which is deemed sufficient as larger constants can be constructed through multiplication, and it is advisable to perform coordinate transformations for extremely large constants. The initial size of the candidate tree is fixed as 5, with an upper limit of 15 during the iterative process. In the tournament selection method, we set the size of a group as 2.

The ratio of the three genetic operators, i.e., crossover, mutation for subtree, and mutation for constant, is set as $0.6:0.35:0.05$ in this paper. These operators generate $80\%$ of the new individles and others are produced by reproduction operator. In the operation of mutation for constant, the parameter $\vartheta$ is chosen as 10. The parameter $\tau_2$ is a small positive number, commonly set within the range $\left( 0,0.1 \right]$. The selection of $\tau_2$ is guided by the criterion of gradually and slowly reducing the node count of the best individual before the $N_{thre}$-th generation. The initial value of $\delta \tau$ is chosen as $\delta \tau=0.08$. We choose the parameter $N_{thre}$ as 50 initially and adjust it finely based on the performance. The parameter $E_{thre}$ is selected according to the magnitude of the system, commonly $0.001 \sim 0.1$ times the initial mean squared error.

\section{Numerical experiments}
\label{NUEXsec}
In this section, we demonstrate the application of our proposed algorithm through two illustrative examples, aiming to discover the underlying governing equations of stochastic dynamical systems excited with both (Gaussian) Brownian noise and (non-Gaussian) L\'evy noise from sample path data. To this end, we select a two-dimensional Maier-Stein system and a three-dimensional chaotic system as representative models. Due to the randomness of the method, we apply the algorithm five times and present the best solution for each term. These examples serve to affirm the efficacy and precision of our methodology in discerning the complex dynamics inherent in such systems.

\subsection{Maier-Stein system}
\label{MSsec}
We now turn our attention to the two-dimensional Maier-Stein stochastic dynamical system, a well-known double-well overdamped model that exhibits intricate dynamical behaviors. This system is described by the following stochastic differential equations \cite{maier1996scaling}:
\begin{equation}\label{MS}
	\begin{aligned}
	& dx_1= (x_1-x_1^3-x_1 x_2^2)dt + \sin \frac{\pi x_1}{2}dB_{1t} + dL_{1t}, \\
	& dx_2= -(1+x_1^2)x_2 dt + 0.5x_2 dB_{2t} + dL_{2t}.
	\end{aligned}
\end{equation}
The drift coefficient, which is a vector field in $\mathbb{R}^2$, is given by $b(x)= [x_1-x_1^3-x_1 x_2^2, -(1+x_1^2)x_2]^T$. The diffusion coefficient is defined by the matrix $a(x)= \left[ \begin{matrix} \sin ^2 \frac{\pi x_1}{2} & 0 \\ 0 & 0.25x_2^2 \end{matrix} \right]$. The L\'evy noise intensity is set as $\sigma_2=1$, and the kernel function of the jump measure is expressed as $W(y)= c(2,\alpha)|y|^{-2-\alpha}$ where $c(2,\alpha)= \frac{\alpha \Gamma((2+\alpha)/2)}{2^{1-\alpha} \pi \Gamma(1-\alpha/2)}$ and $\alpha=1.5$. The time step for simulations, $h$, is set to 0.001. The initial conditions $Z=[z_1,...,z_M]$ are randomly distributed across the domain $[-2,2] \times [-2,2]$ with $M=2 \times 10^8$. This domain is partitioned into a grid of $20 \times 20$ bins to facilitate the inference of drift and diffusion terms. Then the dataset $X$ is generated using Euler scheme, initiated from the points in $Z$. The simulation technique for rotaionally symmetric $\alpha$-stable L\'evy noise can refer to Nolan \cite{Nolan}.

First, we employ the propposed method to discover the L\'evy jump measure. The algorithm is configured with a maximum generation limit of $G_{max}=100$ and a population size of $N_P=500$. Each individual in the initial population comprises 5 candidate functions and it may decrease along with the iteration due to the editting operation. The threshold parameters are fixed as $N_{thre}=20$ and $E_{thre}=10^{-5}$. The error threshold that signals the termination of the algorithm is set to $e_{thre}= 5 \times 10^{-7}$. The parameter $\tau_2$ is assigned as $\tau_2=0.1$. The function set and the terminal set are selected as follows
\begin{equation*}
	\{ +, \times, /, \exp, \ln, \text{squ} \},
\end{equation*}
\begin{equation*}
	\{ 1, c, r \},
\end{equation*}
where $r=|x|$. The sine function has been excluded from the function set, as it is not typically observed in the L\'evy jump measure.

\begin{figure}
	\centering
	\includegraphics[width=11cm]{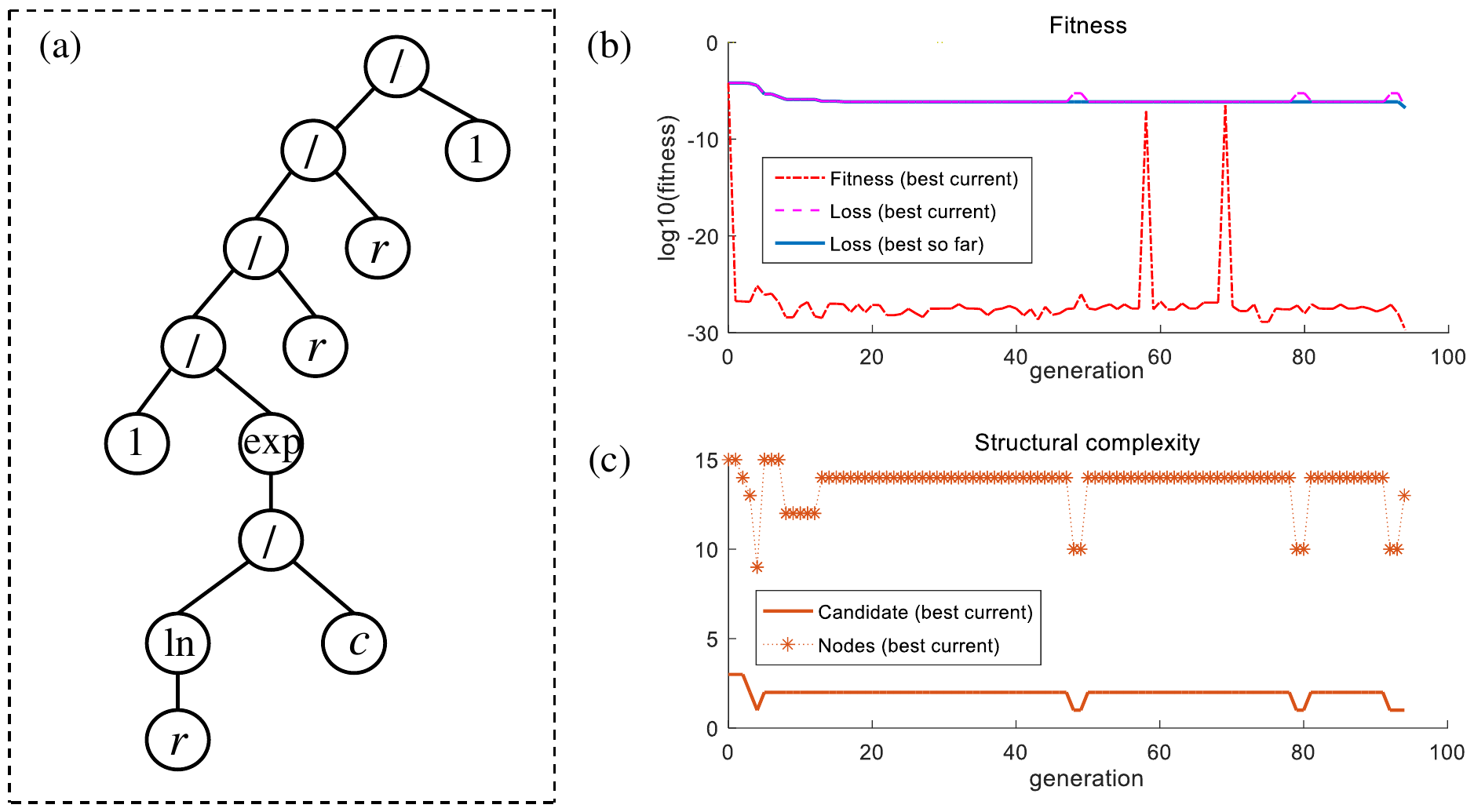}
	\caption{The learned results of L\'evy jump measure of stochastic Maier-Stein system. (a) The optimal individle of the jump measure after performing the algorithm. The constant $c$ in the tree structure is 2.0368. (b) The mean squared loss functions for current best inidividle and the best inidividle so far and the fitness measure during the iterating process. (c) The candidate number and total nodes in the current best inidividle during the iterating process.}
	\label{fig4}
\end{figure}

Fig. \ref{fig4} illustrates the outcomes of our algorithm in discovering the L\'evy jump measure of the stochastic Maier-Stein system. As shown in Fig. \ref{fig4}(b), the mean squared loss after 94 generations reaches a value of $1.7 \times 10^{-7}$, which is below the predefined error threshold $e_{thre}$, leading to the algorithm's termination at this point. It is also noteworthy that the fitness measure has an exceedingly small value, on the order of $10^{-28}$. This phenomenon can be attributed to the random combination of function operators and state variables, which may result in candidate functions with very large magnitudes, thereby significantly increasing $\max{\mathcal{L}}_b$ within the fitness (\ref{fitness}). By the final generation, the optimal solution is distilled to a single candidate function comprising 13 nodes, as seen in Fig. \ref{fig4}(c). The tree structure of this solution is displayed in Fig. \ref{fig4}(a), with an associated sparse learning coefficient of 1.0735. Consequently, the estimated jump measure is expressed as
\begin{equation*}
	1.0735 \times 1/ \exp \left\{ \ln r /2.0368 \right\} /r /r /1 = 1.0735 r^{-2.4910}.
\end{equation*}
Given the known values of $\alpha=1.5$ and $\sigma_2=1$, the integrand in Eq. (\ref{jm123}) is indeed represented by
\begin{equation*}
	2\pi \sigma_2^{-2} rW(\sigma_2^{-1}r) = 1.0755r^{-2.5},
\end{equation*}
which is perfectly consistent with the learned jump measure. This not only confirms the algorithm's ability to identify the type of L\'evy motion as $\alpha$-stable but also enables the extraction of the L\'evy motion parameters $\alpha$ and $\sigma_2$ directly from the data. It is important to emphasize that Fig. \ref{fig4}(a) is not the sole tree structure that can estimate the jump measure. For example, an alternative expression could be $1.0805 \times \exp \left\{ -2.4905 \times \ln r \right\}$.

Second, we utilize the evolutionary algorithm to extract the drift coeffcient of the stochastic Maier-Stein system from data. The algorithm is configured with a maximum generation limit of $G_{max}=100$ and a population size of $N_P=1000$ individles. Each individual in the initial population is endowed with 20 candidate functions. The threshold parameters are fixed as $N_{thre}=50$ and $E_{thre}=0.1$. The error threshold that triggers the termination of the algorithm is defined as $e_{thre}= 0.001$. The parameter $\tau_2$ is allocated a value of $\tau_2=0.02$. The function set and the terminal set are selected as follows
\begin{equation*}
	\{ +, \times, /, \exp, \ln, \sin, \text{squ} \},
\end{equation*}
\begin{equation*}
	\{ 1, c, x_1, x_2\}.
\end{equation*}

\begin{figure}
	\centering
	\includegraphics[width=11cm]{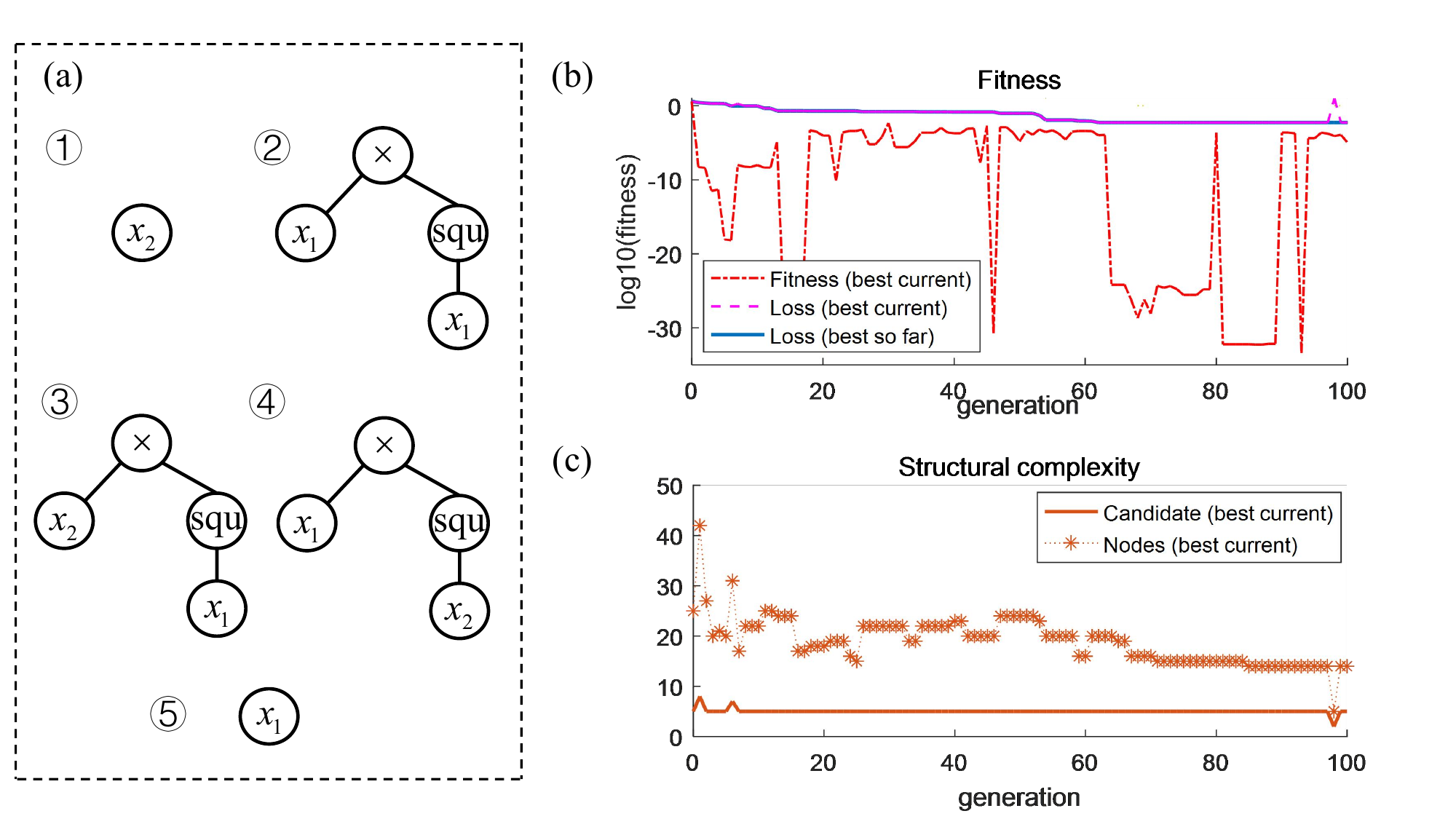}
	\caption{The learned results of drift coefficient of stochastic Maier-Stein system. (a) The optimal individle of the drift coefficient after performing the algorithm. (b) The mean squared loss functions for current best inidividle and the best inidividle so far and the fitness measure during the iterating process. (c) The candidate number and total nodes in the current best inidividle during the iterating process.}
	\label{fig5}
\end{figure}

The learned results of drift coefficient of stochastic Maier-Stein system are depicted in Fig. \ref{fig5}. As observed in Fig. \ref{fig5}(b) and (c), the mean squared error function is reduced to 0.0054 after a complete run of 100 iterations. The final best individual comprises 5 candidate functions, encompassing a total of 14 nodes. The tree structures of these 5 candidates are illustrated in Fig. \ref{fig5}(a), with their corresponding mathematical expressions listed as
\begin{equation*}
	x_2, x_1^3, x_1^2 x_2, x_1 x_2^2, x_1.
\end{equation*}
The sparse regression approach leads to the following coefficients to approximate the drift term
\begin{equation*}
	\left[\begin{matrix}
		0 & -0.9599 & 0 & -0.9800 & 0.8604 \\
		-0.9901 & 0 & -0.9972 & 0 & 0
	\end{matrix}\right]^T.
\end{equation*}
These results closely resemble the ground truth for the drift term, which is given by
\begin{equation*}
	\left[\begin{matrix}
		0 & -1 & 0 & -1 & 1 \\
		-1 & 0 & -1 & 0 & 0
	\end{matrix}\right]^T.
\end{equation*}
This comparison demonstrates that the proposed evolutionary method is capable of successfully inferring the characteristics of polynomial functions directly from data.

Third, we apply the proposed approach to identify the diffusion term from data. The algorithm is set with a maximum generation limit of $G_{max}=100$ and a population size of $N_P=1000$ individles. Each individual in the initial population is endowed with 20 candidate functions. The threshold parameters are set as $N_{thre}=50$ and $E_{thre}=0.005$. The error threshold that triggers the termination of the algorithm is set to $e_{thre}= 0.0001$. The parameter $\tau_2$ is assigned as $\tau_2=0.04$. The function set and the terminal set are selected as follows
\begin{equation*}
	\{ +, \times, /, \exp, \ln, \sin, \text{squ} \},
\end{equation*}
\begin{equation*}
	\{ 1, c, x_1, x_2\}.
\end{equation*}

\begin{figure}
	\centering
	\includegraphics[width=11cm]{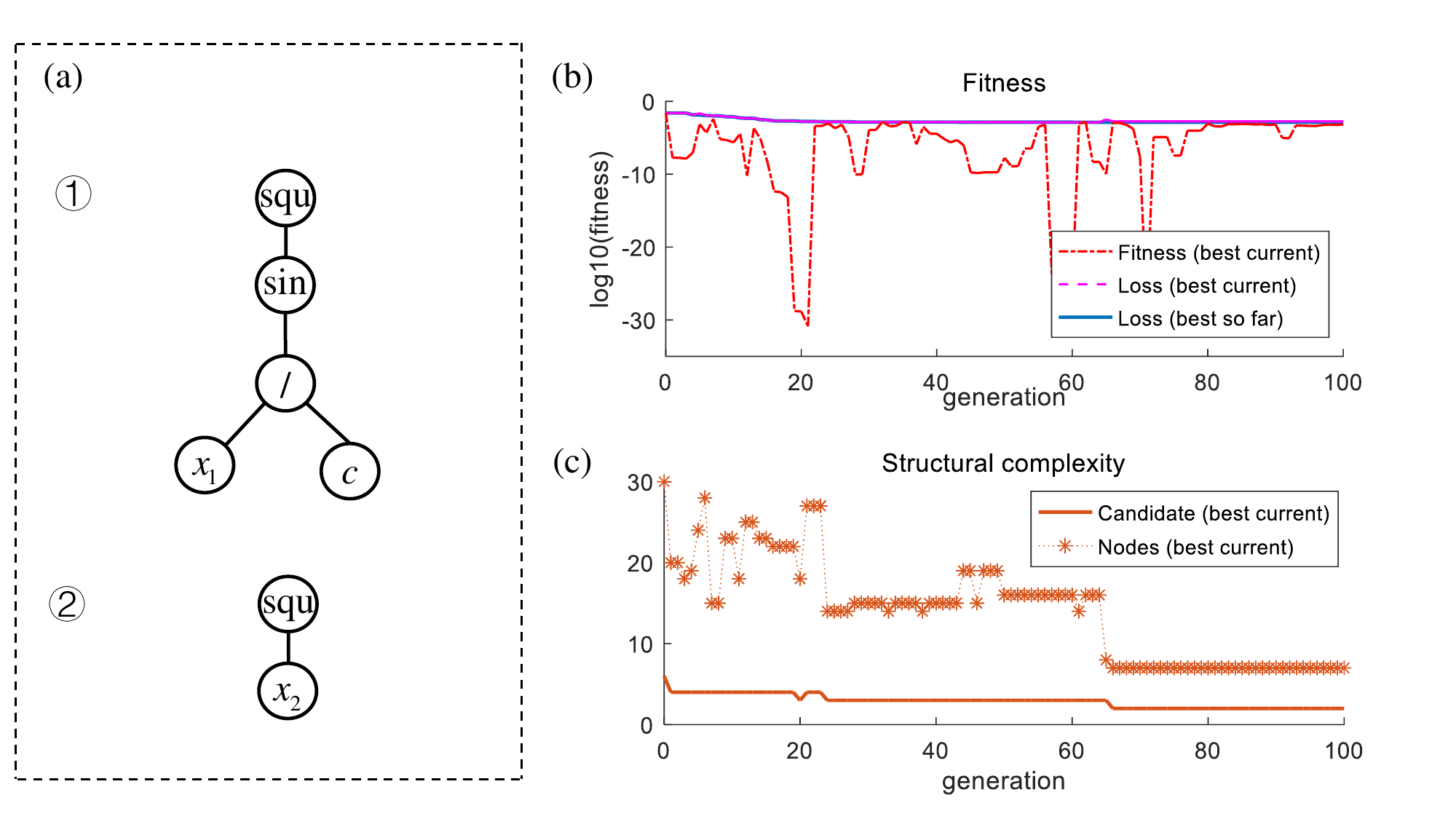}
	\caption{The learned results of diffusion coefficient of stochastic Maier-Stein system. (a) The optimal individle of the diffusion coefficient after performing the algorithm. The constant $c$ in the first candidate has the value -0.6448. (b) The mean squared loss functions for current best inidividle and the best inidividle so far and the fitness measure during the iterating process. (c) The candidate number and total nodes in the current best inidividle during the iterating process.}
	\label{fig6}
\end{figure}

As depicted in Fig. \ref{fig6}(b) and (c), the mean squared error function decreases to 0.0018 after a full complement of 100 generations, and the best individual that emerged comprises just 2 candidate functions, encompassing a total of 7 nodes. Fig. \ref{fig6} delineates the tree structures of these two candidates, with their corresponding mathematical expressions presented as
\begin{equation*}
	\sin ^2 \frac{x_1}{-0.6448} = \sin ^2 1.5509x_1, x_2^2.
\end{equation*}
The sparse regression method results in the following coeffcients for these candidates to estimate the diffusion term
\begin{equation*}
	\left[\begin{matrix}
		0.9314 & 0 & 0 \\
		0 & 0.2462 & 0
	\end{matrix}\right],
\end{equation*}
which agree well with the true results
\begin{equation*}
	a(x)= \left[\begin{matrix}
		\sin ^2 1.5708x_1 & 0 \\
		0 & 0.25x_2^2
	\end{matrix}\right].
\end{equation*}
This concordance confirms that our proposed evolutionary method is capable of extracting complex trigonometric functions with nontrivial constants - a capability that surpasses the reach of standard sparse regression techniques.

\subsection{A three-dimensional chaotic system}
\label{ACSsec}
Let us consider a three-dimensional chaotic dynamical system that incorporates both (Gaussian) Brownian noise and (non-Gaussian) Lévy noise, as described by the following system of stochastic differential equations \cite{xu2014new}:
\begin{equation}\label{ACS}
	\begin{aligned}
		& dx_1= \ln [0.5+\exp(x_2-x_1)]dt + \frac{x_1}{\sqrt{x_1^2 +1}} dB_{1t} + 0.5 dL_{1t}, \\
		& dx_2= x_1 x_3 dt + 0.4x_2 dB_{2t} + 0.5 dL_{2t}, \\
		& dx_3= (1-x_1 x_2) dt + 0.7x_3 dB_{3t} + 0.5 dL_{3t}.
	\end{aligned}
\end{equation}
Here, the terms to be discovered are the drift coefficient, diffusion coefficient, L\'evy noise intensity, and kernel function of jump measure, specified as
\begin{equation}\label{ACS2}
	\begin{aligned}
	    & b(x)= [\ln [0.5+\exp(x_2-x_1)], x_1 x_3, 1-x_1 x_2]^T, \\
	    & a(x)= \left[ \begin{matrix} \frac{x_1^2}{x_1^2 +1} & 0 & 0 \\
	    	 0 & 0.16x_2^2 & 0  \\
    	     0 & 0 & 0.49x_3^2 \end{matrix} \right],  \\
        & \sigma_2=0.5, \ W(y)=c(3,\alpha)|y|^{-3-\alpha},
    \end{aligned}
\end{equation}
where $c(3,\alpha)= \frac{\alpha \Gamma((3+\alpha)/2)}{2^{1-\alpha} \pi^{3/2} \Gamma(1-\alpha/2)}$ and $\alpha=0.5$. The time step for the simulations is set to $h=0.001$ and $M=2 \times 10^8$ initial points are randomly selected within the domain $[-3,3] \times [-3,3] \times [-3,3]$. This region is divided into $9 \times 9 \times 9$ bins for infering drift and diffusion terms. The dataset $X$ is subsequently generated using the Euler scheme with initial points from $Z$.

Similarly, we begin by employing the proposed algorithm to discover the L\'evy jump measure. We set all the hyperparameters as: the maximum generation $G_{max}=100$, the individle number in the population $N_P=1000$, 10 candidate functions in each individle initially, the threshold parameters $N_{thre}=20$ and $E_{thre}=10^{-4}$, the error threshold $e_{thre}= 5 \times 10^{-7}$, and $\tau_2=0.1$. The function set and the terminal set are selected as follows
\begin{equation*}
	\{ +, \times, /, \exp, \ln, \text{squ} \},
\end{equation*}
\begin{equation*}
	\{ 1, c, r \}.
\end{equation*}

\begin{figure}
	\centering
	\includegraphics[width=11cm]{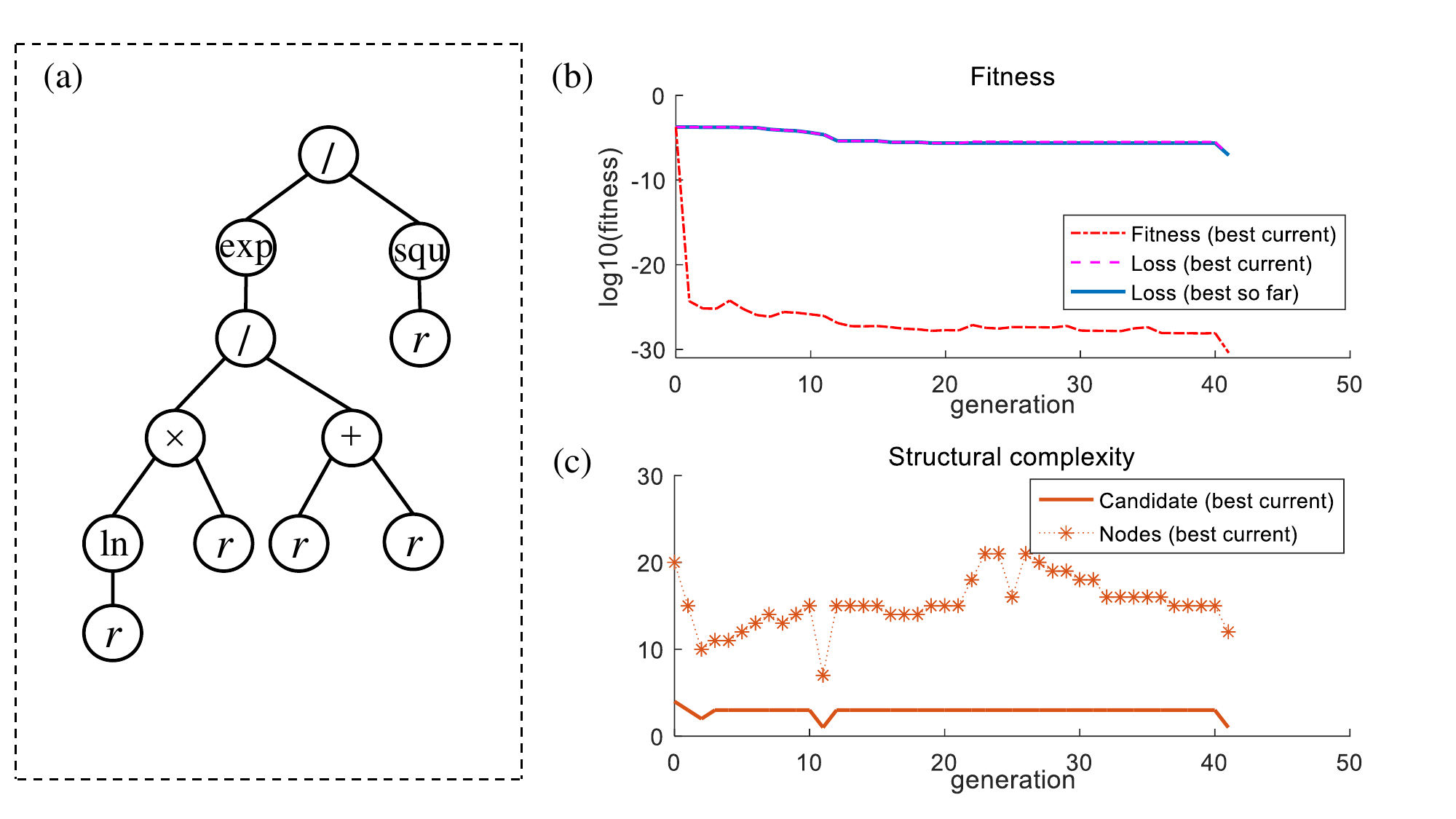}
	\caption{The learned results of L\'evy jump measure of three-dimensional chaotic system. (a) The optimal individle of the jump measure after performing the algorithm. (b) The mean squared loss functions for current best inidividle and the best inidividle so far and the fitness measure during the iterating process. (c) The candidate number and total nodes in the current best inidividle during the iterating process.}
	\label{fig7}
\end{figure}

Fig. \ref{fig7} presents the outcomes of our algorithm in discerning the L\'evy jump measure of the three-dimensional stochastic chaotic system. As shown in Fig. \ref{fig7}(b), the mean squared loss reaches a value of $8.6 \times 10^{-8}$ by the 41st generation, less than the error threshold $e_{thre}$, thus the algorithm terminates therein. In the final generation, the optimal solution is distilled to a single candidate function with 12 nodes, as seen in Fig. \ref{fig7}(c). The tree structure of this solution is exhibited in Fig. \ref{fig7}(a), with an associated sparse learning coefficient of 0.4250. Then the estimated jump measure has the following expression
\begin{equation*}
	0.4250 \exp \left\{ \ln r \times r /(r+r) \right\} /r^2 = 0.4250 r^{-1.5}.
\end{equation*}
In view of $\alpha=0.5$ and $\sigma_2=0.5$, the integrand in Eq. (\ref{jm123}) is indeed given by
\begin{equation*}
	4\pi \sigma_2^{-3} r^2 W(\sigma_2^{-1}r) = 0.4231r^{-1.5},
\end{equation*}
which is perfectly consistent with the learned jump measure.

Subsequently, we use the proposed approach to learn the drift coeffcient of the chaotic system from data. When dealing with complex or high-dimensional systems, it is often advantageous to execute the algorithm multiple times to separately identify the components of the drift vector. In this instance, we segment the components of the drift vector into two groups: the first component and the last two components. For the first group, the algorithm is configured with the following hyperparameters: the maximum generation limit $G_{max}=200$, the population size $N_P=1000$, 20 candidate functions in each individle, two threshold parameters $N_{thre}=50$ and $E_{thre}=0.5$, the error threshold $e_{thre}= 0.001$, and $\tau_2=0.05$. For the second group, the algorithm is set with the following hyperparameters: the maximum generation limit $G_{max}=100$, the population size $N_P=1000$, 20 candidate functions in each individle, two threshold parameters $N_{thre}=50$ and $E_{thre}=0.5$, the error threshold $e_{thre}= 0.001$, and $\tau_2=0.09$. The function set and the terminal set for both cases are selected as follows
\begin{equation*}
	\{ +, \times, /, \exp, \ln, \text{squ} \},
\end{equation*}
\begin{equation*}
	\{ 1, c, x_1, x_2, x_3\}.
\end{equation*}
In certain complex scenarios, the inclusion of the sine function may increase the likelihood of the algorithm becoming trapped in local minima prematurely. Consequently, for the identification of the drift term in this system, we have elected to exclude the sine function from the function set.

\begin{figure}
	\centering
	\includegraphics[width=11cm]{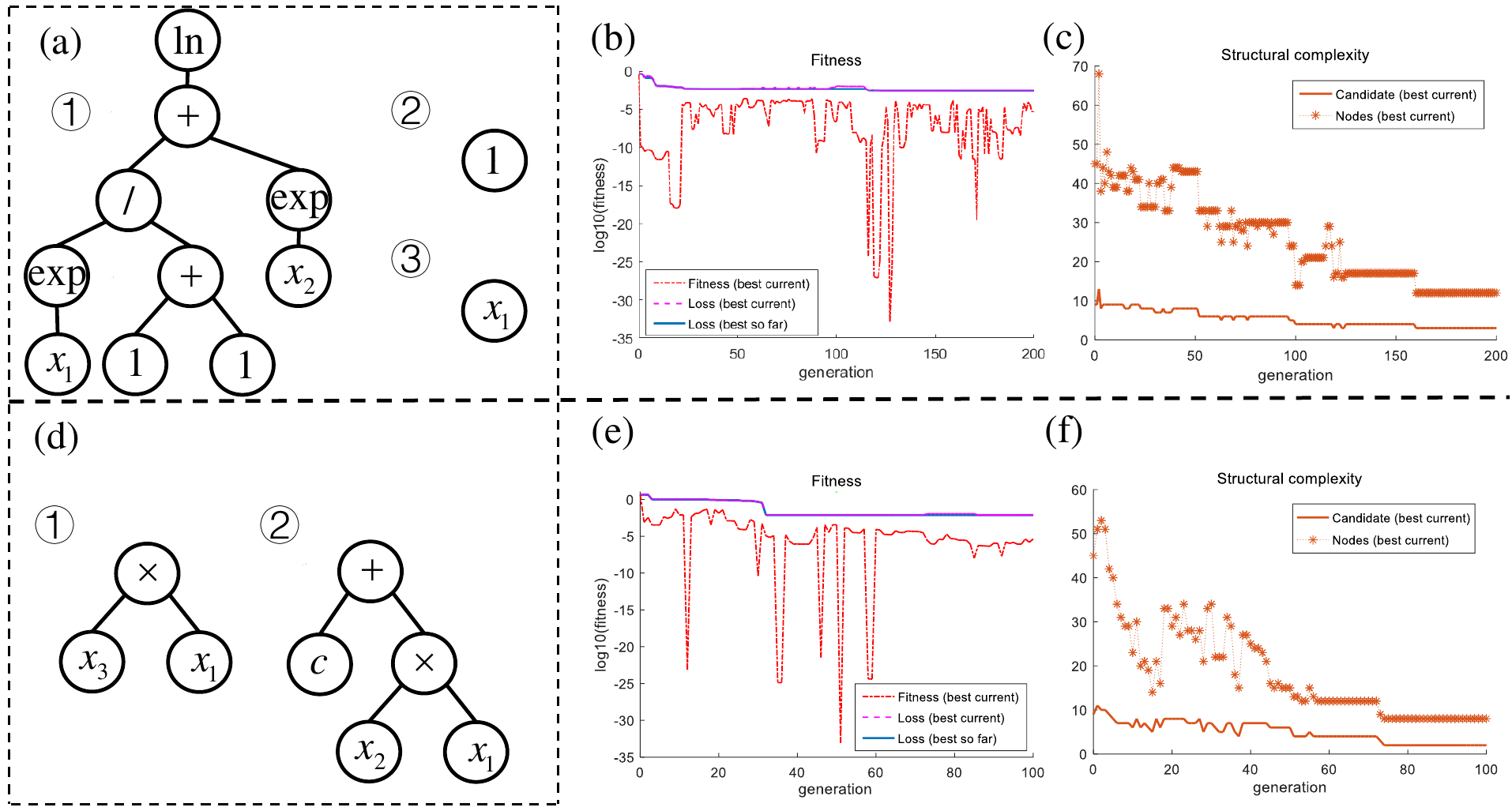}
	\caption{The learned results of drift coefficient of three-dimensional chaotic system. (a,d) The optimal individle of the drift coefficient after performing the algorithm. The constant $c$ in (d) has the value -1.0082. (b,e) The mean squared loss functions for current best inidividle and the best inidividle so far and the fitness measure during the iterating process. (c,f) The candidate number and total nodes in the current best inidividle during the iterating process. (a,b,c) The first component of the drift coefficient. (d,e,f) The last two components of the drift coefficient.}
	\label{fig8}
\end{figure}

The two rows of Fig. \ref{fig8} exhibit the learned results of the two training processes of the algorithm. The mean squared error functions are reduced to 0.0029 and 0.0078, respectively. According to the optimal individles in Fig. \ref{fig8}(a,d) and their sparse coefficients, we obtain the estimted drift term as
\begin{equation*}
	\begin{aligned}
	\tilde{b}(x) &= \left[\begin{matrix}
		0.9955 \ln \left[ \exp(x_1) /(1+1) +\exp(x_2) \right] +0 \times 1 -0.9947x_1 \\
		0.9980x_1 \times x_3  \\
		-0.9987 (-1.0084+x_1 \times x_2) 
	\end{matrix}\right] \\
    &= \left[\begin{matrix}
    	0.9955 \ln \left[ 0.5\exp(x_1) +\exp(x_2) \right] -0.9947x_1 \\
    	0.9980x_1 x_3  \\
    	1.0071-0.9987x_1 x_2
    \end{matrix}\right],
    \end{aligned}
\end{equation*}
which agrees well with the true drift vector
\begin{equation*}
	b(x)= [\ln [0.5 \exp(x_1)+\exp(x_2)] -x_1, x_1 x_3, 1-x_1 x_2]^T.
\end{equation*}

Finally, we utilize the proposed approach to extract the diffusion term from data. The algorithm is configured with the following hyperparameters: the maximum generation limit $G_{max}=200$, the population size $N_P=1000$, 20 candidate functions in each individle, two threshold parameters $N_{thre}=50$ and $E_{thre}=0.05$, the error threshold $e_{thre}= 0.0001$, and $\tau_2=0.05$. The function set and the terminal set are selected as follows
\begin{equation*}
	\{ +, \times, /, \exp, \ln, \text{squ} \},
\end{equation*}
\begin{equation*}
	\{ 1, c, x_1, x_2, x_3\}.
\end{equation*}

\begin{figure}
	\centering
	\includegraphics[width=11cm]{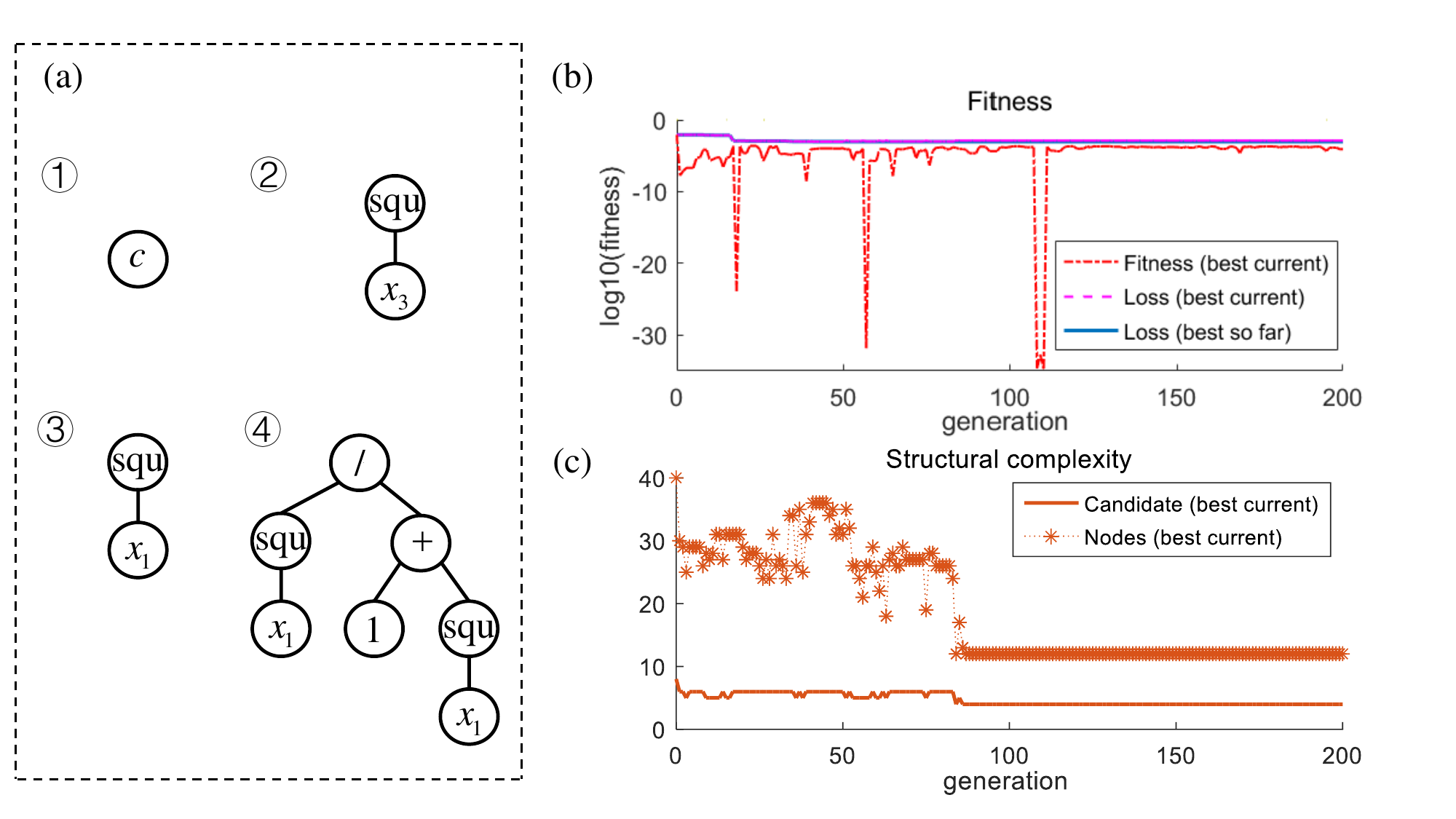}
	\caption{The learned results of diffusion coefficient of three-dimensional chaotic system. (a) The optimal individle of the diffusion coefficient after performing the algorithm. The constant $c$ in the first candidate has the value -0.5105. (b) The mean squared loss functions for current best inidividle and the best inidividle so far and the fitness measure during the iterating process. (c) The candidate number and total nodes in the current best inidividle during the iterating process.}
	\label{fig9}
\end{figure}

As shown in Fig. \ref{fig9}(b) and (c), the mean squared error function decreases to 0.0013 after 200 generations, and the final best individual comprises 4 candidate functions, encompassing a total of 12 nodes. Fig. \ref{fig9} describes the tree structures of these four candidates, with associated mathematical expressions presented as
\begin{equation*}
	c, x_3^2, x_1^2, x_1^2 / (1+x_1^2).
\end{equation*}
The sparse regression method results in the following coeffcients for these candidates to estimate the diffusion term
\begin{equation*}
	\left[\begin{matrix}
		0 & 0 & 0 & 0 & 0 & 0 \\
		0 & 0 & 0.4858 & 0 & 0 & 0 \\
		0 & 0.1563 & 0 & 0 & 0 & 0 \\
		0.9447 & 0 & 0 & 0 & 0 & 0
	\end{matrix}\right],
\end{equation*}
which agree well with the true results
\begin{equation*}
	a(x)= \left[\begin{matrix} \frac{x_1^2}{x_1^2 +1} & 0 & 0 \\
		0 & 0.16x_2^2 & 0  \\
		0 & 0 & 0.49x_3^2 \end{matrix}\right].
\end{equation*}
This accordance shows that the proposed method is able to learn complex rational functions from data.

\section{Conclusion}
\label{CONCsec}
In this research, we introduce the evolutionary symbolic sparse regression (ESSR) method, a data-driven approach devoted to discovering the governing equations of stochastic dynamical systems that incorporate (Gaussian) Brownian noise and (non-Gaussian) L\'evy noise. Our method integrates nonlocal Kramers-Moyal formulas, genetic programming, and sparse regression to extract meaningful insights from data. Specifically, the genetic programming is tasked with generating a spectrum of candidate functions, the sparse regression technique aims at learning the coefficients associated with these candidates, and the nonlocal Kramers-Moyal formulas serve as the foundation for constructing the fitness measure in genetic programming and the loss function in sparse regression.

Through two illustrative examples, we have demonstrated the efficacy and accuracy of our method. It has showcased its capability in extracting the systems with complex functional forms, encompassing trigonometric functions, exponential functions, logrithmic functions, rational functions and their combinations. The ESSR method's key benefits lie in its reduced reliance on prior knowledge when compared to traditional sparse regression techniques, and its superior interpretability over neural networks, as it can explicitly reveal the governing equations of the system. This approach offers a robust tool for analyzing and forecasting the behavior of complex stochastic systems, with broad applications across scientific and engineering disciplines.

While the ESSR method has some distinct advantages for identifying the governing equations of non-Gaussian stochastic dynamical systems, it also faces several challenges that warrant attention. Firstly, extending the application of this method to high-dimensional systems is not trivial and requires further research and development. Secondly, the method's efficacy hinges on the availability of a substantial amount of data, which can be difficult to obtain in real-world scenarios. Thirdly, learning from data with noise that significantly overpowers the drift term is challenging, as it necessitates an even more extensive dataset for effective preprocessing. Fourthly, the method involves a multitude of hyperparameters that necessitate meticulous tuning to ensure optimal performance. Addressing these challenges will be the focus of our future work. By overcoming these obstacles, we aim at further refining our ESSR method, thereby expanding its applicability and enhancing its potential as a powerful tool in the field of data-driven system identification.

\section*{Appendix. L\'evy motion}
\label{Levy}
Let $L_{t}$ be a stochastic process in $\mathbb{R}^n$ defined on a probability space $(\Omega, \mathcal{F}, P)$. We refer to $L_{t}$ as a L\'evy motion if it satisfies the following conditions: \\
(1) $L_{0}=0$, a.s.;\\
(2) $L_{t}$ has independent and stationary increments;\\
(3) $L_{t}$ is stochastically continuous, i.e., for all $\delta>0$ and for all $s\geq0$
$$
\lim\limits_{t\to s}P(|L_{t}-L_{s}|>\delta)=0.
$$

The characteristic function of a L\'evy motion $(L_{t},t\geq0)$ is given by the L\'evy-Khinchine formula, which for each $t\geq0$, $u\in\mathbb{R}^n$, takes the form:
\begin{align*}
	& \mathbb{E} \text{exp} \left[ i\langle u,L_t \rangle \right] = \text{exp} \left[ t \eta \left( u \right) \right],\\
	& \eta \left( u \right) = i\langle b,u \rangle -\frac{1}{2} \langle u,Au \rangle  + \int_{\mathbb{R}^n \backslash\{0\}} \left[ e^{i \langle u,y \rangle}-1-i \langle u,y \rangle I_{ \left\{\Vert y\Vert<1 \right\}}(y) \right]\, \nu(dy),
\end{align*}
where $(b,A,\nu)$ is the characteristic triplet of the L\'evy motion $L_{t}$. In many cases, we focus on the pure jump case $\left( 0,0,\nu \right)$.

A rotationally symmetric $\alpha$-stable L\'evy motion is notable type of L\'evy process characterized by its stable random vector. The jump measure for this process is given by
\begin{align}
	\label{jumpm}
	\nu \left( dy \right) = c \left( n, \alpha \right) {\left| y \right|} ^ {-n-\alpha} dy
\end{align}
for $y \in \mathbb{R}^{n} \backslash \{0\}$ with the intensity constant $c \left( n, \alpha \right)$ defined as
\begin{align}
	\label{jumpc}
	c \left( n, \alpha \right) = \frac {\alpha \Gamma \left( \left( n+\alpha \right)/2 \right)} {2^{1-\alpha} {\pi}^{n/2} \Gamma \left( 1-\alpha/2 \right)}.
\end{align}
This process exhibits larger jumps with lower frequencies for smaller values of $\alpha$ ($0<\alpha <1$), and smaller jumps with higher frequencies for larger values of $\alpha$ ($1<\alpha <2$). The special case $\alpha=2$ corresponds to (Gaussian) Brownian motion. For further insights into L\'evy motion, the reader is directed to the referenced works \cite{Duan2015, Applebaum}.

\section*{Acknowledgement}
The authors would like to thank Prof. Ehsan Askari for helpful discussions and Prof. Sara Silva for GPLAB - A Genetic Programming Toolbox for MATLAB. The authors acknowledge support from the National Natural Science Foundation of China Grant No. 12302035 and the Natural Science Foundation of Jiangsu Province Grant No. BK20220917. This work was also partly supported by the National Natural Science Foundation of China grant 12141107 and  the Guangdong-Dongguan Joint Research Grant 2023A1515140016.

\section*{Competing Interests}


The authors declare that they have no conflict of interest.

\section*{Data Availability}
The data that support the findings of this study are available from the corresponding author upon reasonable request.


%
%


\bibliographystyle{abbrv}
\bibliography{manuscript}

%
%


\end{document}